\documentclass[twoside]{article}
\usepackage[accepted]{aistats2026}

\usepackage[utf8]{inputenc} %
\usepackage[T1]{fontenc}    %

\usepackage{amsmath}
\usepackage{amssymb}
\usepackage{mathtools}
\usepackage{amsthm}
\usepackage{enumitem}
\usepackage{algorithm}
\usepackage{algpseudocode}
\usepackage{microtype}
\usepackage{graphicx}
\usepackage{subfigure}
\usepackage{booktabs}

\usepackage[authoryear,round]{natbib}
\usepackage[colorlinks,linkcolor={red!80!black},citecolor={blue},allbordercolors={1 1 1},urlcolor={blue!80!black},hypertexnames=false]{hyperref}
\usepackage[capitalize,noabbrev]{cleveref}
\usepackage{setspace}
\usepackage[dvipsnames]{xcolor}
\usepackage{tikz}
\usetikzlibrary{positioning}
\usepackage{svg}

\theoremstyle{plain}
\newtheorem{theorem}{Theorem}[section]

\newtheorem{lemma}[theorem]{Lemma}
\newtheorem{corollary}[theorem]{Corollary}
\theoremstyle{definition}
\newtheorem{definition}[theorem]{Definition}

\theoremstyle{remark}

\newcommand{\Ftw}[1]{\mathcal{F}_{t,w}(#1)}

\newcommand{\Bcal}{\mathcal{B}}
\newcommand{\mathbbE}{\mathbb{E}}

\newcommand{\textBLR}{\text{BLR}}
\newcommand{\losstw}{\mathcal{F}_{t,w}}
\newcommand{\inOBJ}{L_t^{in}}
\newcommand{\outOBJ}{{L_{t}^{out}}}
\newcommand{\pointWinOBJ}{\ell_{in}}
\newcommand{\pointWoutOBJ}{\ell_{out}}
\newcommand{\dualOBJ}{L_{adj}}
\newcommand{\innerpred}{h}
\newcommand{\adjoint}{a}
\newcommand{\inNNparam}{\theta}
\newcommand{\adjNNparam}{\xi}
\newcommand{\outvar}{{\boldsymbol{\lambda}}}
\newcommand{\outgradt}{\nabla\mathcal{F}_{t}}
\newcommand{\outgradtw}{\nabla\losstw}
\newcommand{\outgradti}{\nabla\mathcal{F}_{t-i}}

\newcommand{\outgradhatt}{\widehat{\nabla\mathcal{F}}_{t}}
\newcommand{\outgradhatti}{\widehat{\nabla\mathcal{F}}_{t-i}}
\newcommand{\outgradtildt}{\widetilde{\nabla\mathcal{F}}_{t,w}}

\newcommand{\empinOBJ}{\hat{L}_{in}}
\newcommand{\empoutOBJ}{\hat{L}_{out}}
\newcommand{\empadjOBJ}{\hat{L}_{adj}}

\newcommand{\Hessian}{C}
\newcommand{\crossDeriv}{B}

\newcommand{\nbItersInnerSol}{M}

\newcommand{\nbItersAdjSol}{K}
\newcommand{\outDataset}{\mathcal{D}_{out}}
\newcommand{\inDataset}{\mathcal{D}_{in}}
\newcommand{\outBatch}{\mathcal{B}_{out}}
\newcommand{\inBatch}{\mathcal{B}_{in}}

\newcommand{\bigO}{\mathcal{O}}
\DeclareMathOperator*{\argmin}{arg\,min}
\newcommand{\Real}{\mathbb{R}}
\newcommand{\Hcal}{\mathcal{H}}

\newcommand{\parens}[1]{\left( #1 \right)}
\newcommand{\brackets}[1]{\left[ #1 \right]}
\newcommand{\verts}[1]{\left\lvert #1 \right\rvert}
\newcommand{\Verts}[1]{\left\lVert #1 \right\rVert}

\algnewcommand{\IIf}[1]{\State\algorithmicif\ #1\ \algorithmicthen}
\algnewcommand{\EndIIf}{\unskip\ \algorithmicend\ \algorithmicif}

\newlist{assumplist}{enumerate}{1}
\setlist[assumplist]{label=(\textbf{\Alph*})}
\Crefname{assumplisti}{Assumption}{Assumptions}

\newlist{assumplist2}{enumerate}{1}
\setlist[assumplist2]{label=(\textbf{\alph*})}
\Crefname{assumplist2i}{Assumption}{Assumptions}

\definecolor{darkred}{rgb}{0.55, 0.0, 0.0}

\begin{document}
\twocolumn[

\aistatstitle{Non-Stationary Functional Bilevel Optimization}

\aistatsauthor{
Jason Bohne\textsuperscript{1}\footnotemark[1] \And
Ieva Petrulionyte\textsuperscript{2}\footnotemark[1] \And
Michael Arbel\textsuperscript{2} \And
Julien Mairal\textsuperscript{2} \And
Pawe\l\ Polak\textsuperscript{1}
}

\aistatsaddress{
\textsuperscript{1}Applied Mathematics and Statistics, Stony Brook University, Stony Brook, NY, USA\\
\textsuperscript{2}Inria, CNRS, Grenoble INP, LJK, Université Grenoble Alpes\\
}

]
\footnotetext[1]{Equal contribution.}

\begin{abstract}
Functional bilevel optimization (FBO) provides a powerful framework for hierarchical learning in function spaces, yet current methods are limited to static offline settings and perform suboptimally in online, non-stationary scenarios. We propose \textbf{SmoothFBO}, the first algorithm for non-stationary FBO with both theoretical guarantees and practical scalability. SmoothFBO introduces a time-smoothed stochastic hypergradient estimator that reduces variance through a window parameter, enabling stable outer-loop updates with sublinear regret. Importantly, the classical parametric bilevel case is a special reduction of our framework, making SmoothFBO a natural extension to online, non-stationary settings. Empirically, SmoothFBO consistently outperforms existing FBO methods in non-stationary hyperparameter optimization and model-based reinforcement learning, demonstrating its practical effectiveness. Together, these results establish SmoothFBO as a general, theoretically grounded, and practically viable foundation for bilevel optimization in online, non-stationary scenarios.
\end{abstract}

\section{Introduction}

\par Bilevel optimization has emerged as a powerful paradigm for solving complex nested optimization problems in various domains. Initially employed in model selection \citep{bennett2006model} and sparse feature learning \citep{mairal2011task}, this approach subsequently emerged as an efficient alternative to grid search for hyperparameter optimization tuning \citep{Feurer:2019,Lorraine2019OptimizingMO,Franceschi2017ForwardAR}. More recently, the application domain of bilevel optimization has expanded considerably to encompass meta-learning \citep{DBLP:conf/iclr/BertinettoHTV19}, 
auxiliary task learning \citep{navon2021auxiliary}, inverse problems \citep{Holler_2018}, invariant risk minimization~\citep{arjovsky2019invariant,ahuja2020invariant} and reinforcement learning \citep{Hong:2020a,Liu:2021f,Nikishin:2022}. Traditional bilevel optimization approaches typically operate in parameter spaces, where the inner and outer optimization problems are formulated over finite-dimensional vectors. Recently, \citet{petrulionyte2024functional} introduced Functional Bilevel Optimization (FBO), which extends this framework to function spaces, allowing for an expressive theoretical framework and a novel algorithmic approach.

Despite the success of bilevel optimization methods, current algorithms predominantly address static environments where loss functions are expectations over stationary data distributions. However, numerous real-world applications involve dynamic environments where data distributions change over time \citep{besbes2015non}. In reinforcement learning, for instance, environment dynamics may evolve over time, requiring agents to continuously adapt their policies \citep{besbes2015non,padakandla2020reinforcement}. Similarly, in online learning, the patterns in data often change systematically over time \citep{tarzanagh2024online}. These challenges motivate the development of bilevel optimization methods that can efficiently handle such non-stationary settings. While limited research has explored parametric bilevel approaches in non-stationary environments \citep{bohne2024online,tarzanagh2024online}, a comprehensive functional perspective remains undeveloped for such applications, representing a significant gap in the literature.

In this paper, we address this gap by introducing Non-Stationary Functional Bilevel Optimization (\ref{def:FBO}), a framework that extends functional bilevel optimization to time-varying settings. Formally, we aim to solve $\forall t \in [1,T]$:
\begin{equation}\tag{NS-FBO}\label{def:FBO}
\begin{aligned}
	\min_{\outvar \in \Lambda}\ \mathcal{F}_t(\outvar) 
    &:= \outOBJ \bigl(\outvar, \innerpred^\star_{t,\outvar}\bigr) \\
    \text{s.t.}\quad  
    \innerpred^\star_{t,\outvar} 
    &=  \argmin_{\innerpred\in\Hcal}\ \inOBJ\bigl(\outvar, \innerpred\bigr).
\end{aligned}
\end{equation}

\vspace{-0.2cm}
Where we have a sequence of time-varying loss functions $(\outOBJ,\inOBJ)$ for $t=1,\ldots, T$ due to time-varying data distributions $\mathbb{P}_t,\mathbb{Q}_t$. With this definition, a special case of an \ref{def:FBO} problem is the following stochastic non-stationary bilevel optimization problem defined $\forall t\in[1,T]$,

\begin{equation}\label{eq:stoc_special_case}
\begin{aligned}
	\min_{\outvar \in \Lambda}\ 
	\outOBJ\bigl(\outvar,\innerpred^\star_{t,\outvar}\bigr) 
    &:= \mathbb{E}_{x,y\sim\mathbb{P}_t}\!\left[
    \pointWoutOBJ\!\left(\outvar,\innerpred_{t,\outvar}^\star(x),x,y\right)\right] \\
    \text{s.t.}\quad  
    \innerpred^\star_{t,\outvar} 
    &=  \argmin_{\innerpred\in \Hcal}\ 
	\mathbb{E}_{x,y\sim \mathbb{Q}_t}\!\left[
    \pointWinOBJ\!\left(\outvar,\innerpred(x),x,y\right)\right].
\end{aligned}
\end{equation}

where $\mathcal{F}_t$ represents the outer objective function at time $t$, which depends on the solution to the inner problem $\innerpred^\star_{t,\outvar}$, and $\pointWinOBJ, \pointWoutOBJ$ are point-wise losses. Differently from classical parametric approaches, the inner variable $\innerpred_{t,\outvar}$ is a function living in some function space $\Hcal$. The inner and outer objectives involve expectations over potentially different data distributions $\mathbb{P}_t$ and $\mathbb{Q}_t$ that evolve over time. We denote $\Omega = \mathbb{P}_t \times \mathbb{Q}_t$ as the joint distribution  from data samples  $(x_t, y_t) \sim \mathbb{P}_t$ from the outer objective  and $(x_t, y_t) \sim \mathbb{Q}_t$ from the inner objective.

Our work bridges the gap between functional bilevel optimization and online learning, enabling efficient optimization in non-stationary environments. The contributions of this paper are summarized as follows:

\begin{enumerate}
    \item We formulate the Non-Stationary Functional Bilevel Optimization (\ref{def:FBO}) problem for stochastic settings with time-varying data distributions.
    \item We develop \textit{SmoothFBO}, an efficient algorithm that incorporates time-smoothing techniques to handle temporal dependencies and reduce variance in gradient estimates.
    \item We provide theoretical convergence guarantees for our proposed algorithm.
    \item We demonstrate the practical efficacy of our approach on a controlled synthetic non-stationary regression task and on model-based reinforcement learning in non-stationary environments.
\end{enumerate}

The remainder of this paper is organized as follows: \textit{Section 2} presents the preliminaries from functional and online bilevel optimization literature; \textit{Sections 3 and 4} provide our proposed methods and the theoretical analysis of its convergence properties; \textit{Section 5} presents experiments on a synthetic non-stationary regression and on model-based reinforcement learning.

\section{Related Work}

\paragraph{Bilevel optimization and hypergradients.}
Bilevel optimization is a standard tool for hyperparameter tuning and meta-learning. Most practical methods rely on either differentiating through an inner solver (``unrolling'') or using implicit differentiation to avoid storing the full trajectory \citep{Franceschi2017ForwardAR,Lorraine2019OptimizingMO,Shaban:2019,arbel2022amortized}. Our work follows this line, but targets a setting where the objectives drift over time and where the inner problem is posed in a function space rather than in a finite-dimensional parameterization.

\paragraph{Functional bilevel optimization.}
Functional Bilevel Optimization represents a paradigm shift from traditional parameter-centric approaches to a function-space perspective for bilevel optimization problems. This framework, introduced by \citet{petrulionyte2024functional}, effectively addresses the ambiguity challenges that emerge when employing deep neural networks in bilevel optimization settings. The functional perspective provides a theoretical framework that accurately describes the actual techniques used by machine learning practitioners and derives a novel functional implicit differentiation rule.
In this work we keep the same functional viewpoint, but move from an offline regime to a non-stationary one. Algorithmically, we introduce a time-smoothed stochastic estimator of the functional hypergradient and show how the window controls the stability--adaptation tradeoff.

\paragraph{Online / non-stationary bilevel optimization.}
Recent work has started to analyze bilevel learning with time-varying objectives in the parametric setting, typically under smoothness/strong-convexity assumptions and with regret guarantees. In particular, \citet{lin2023non} study nonconvex bilevel problems with time-varying objectives, and more recent online bilevel optimization methods use window-averaged hypergradients to reduce variance and improve tracking \citep{bohne2024online,tarzanagh2024online}. For instance, \citet{bohne2024online} showed that averaging past gradients mitigates the impact of stochastic noise, enabling better tracking of slowly changing optima in parametric bilevel problems. However, these approaches are limited to finite-dimensional parameter spaces, which may not capture the expressive power of function spaces required for complex tasks like those in  reinforcement learning. \textit{SmoothFBO} can be seen as a functional generalization of this idea: we apply time-smoothing at the level of functional hypergradients (estimated from data), which lets us treat parametric bilevel optimization as a special case while retaining the expressivity of function spaces needed for applications such as reinforcement learning.

\paragraph{Online / non-stationary multi-level optimization.}
Our regret analysis also connects to the broader online optimization literature, where sublinear dynamic regret typically requires controlling the temporal variation of the losses (``gradual variation'') \citep{besbes2015non,chiang2012online,yang2016tracking}. A related viewpoint appears in time-varying multi-objective optimization, where one seeks guarantees for multiple competing objectives as they drift in time \citep{shafiei2025tradeoff}. While our setting is bilevel (rather than multi-level), we share the same goal of making the stability-performance tradeoff transparent, here via an explicit window parameter. Finally, multi-level \emph{convex} problems have been studied through monotone operator theory and fixed-point arguments. \citet{shafiei2024trilevel} propose first-order fixed-point algorithms for nested convex programs and analyze convergence rates under monotonicity assumptions. This is complementary to our work: we focus on stochastic, non-stationary bilevel learning, and we emphasize function-space modeling, whereas monotone-operator analyses typically target deterministic, finite-dimensional convex formulations.

\section{Preliminaries}
\subsection{Functional Bilevel Optimization}
\label{sec:FBO}

Here we describe how Functional Bilevel Optimization (FBO) can be effectively used to compute the hypergradient for the non-stationary bilevel problem introduced in the previous section. The non-stationary functional bilevel problem (\ref{def:FBO}) involves an optimal prediction function $\innerpred^\star_{t,\outvar}$ for each value of the outer-level parameter $\outvar$. Solving \ref{def:FBO} by using a first-order method then requires characterizing the implicit dependence of $\innerpred^\star_{t,\outvar}$ on the outer-level parameter~$\outvar$ to evaluate the hypergradient $\outgradt$ in~$\Real^d$. Using functional implicit differentiation and the adjoint sensitivity method from \citet{petrulionyte2024functional}, under differentiability and standard optimization assumptions detailed in \cref{appx:SmoothFBO_assumptions}, the functional hypergradient $\outgradt$ is given by:
\begin{equation}\label{eq:hyper_grad}
\begin{aligned}
	\outgradt(\outvar) 
    &= \partial_\outvar \outOBJ(\outvar, \innerpred^\star_{\outvar}) 
    + \crossDeriv_{\outvar}\, \adjoint_{t,\outvar}^\star, \\
    \text{with }\quad
   \crossDeriv_{\outvar} &:= \partial_{\outvar,\innerpred}\inOBJ(\outvar,\innerpred^\star_{t,\outvar}).
\end{aligned}
\end{equation}

with $\adjoint^{\star}_{t,\outvar}:=- \Hessian_{\outvar}^{-1} d_{\outvar}$ an element of $\Hcal$ that minimizes the quadratic objective: 
\begin{equation}\label{eq:adjoint_objective}
\begin{aligned}
\adjoint_{\outvar}^\star 
   &= \arg\min_{\adjoint\in \Hcal}\ 
   \dualOBJ(\outvar, \adjoint) \\
   &:= \tfrac{1}{2}\ \langle \adjoint, \Hessian_{\outvar} \adjoint \rangle_{\Hcal} 
   + \langle \adjoint, d_{\outvar}\rangle_{\Hcal}, \\
   \text{with }\quad
   \Hessian_{\outvar} 
   &:= \partial^2_{\innerpred} \inOBJ(\outvar,\innerpred^\star_{t,\outvar}).
\end{aligned}
\end{equation}

FBO is a class of practical algorithms designed to estimate the functional hypergradient $\outgradt(\outvar)$ in the stationary context. Our goal is to generalize this class of algorithms to non-stationary environments when both the outer and inner losses are expectations on time-varying probability distributions. 

\subsection{Online Bilevel Optimization}\label{sec:online_bo}
Online Bilevel Optimization (OBO) extends the bilevel optimization framework to dynamic, non-stationary environments where objectives evolve over time due to shifting data distributions or environmental changes. Unlike traditional bilevel optimization, which relies on static datasets, OBO involves sequential learning in response to streaming data, making it critical for applications like online reinforcement learning \citep{padakandla2020reinforcement}. In such settings, the inner and outer objectives, defined over time-varying distributions \(\mathbb{P}_t\) and \(\mathbb{Q}_t\), require algorithms to adaptively track drifting optima while maintaining optimization stability. 

To ensure meaningful regret bounds in non-stationary environments, such as the stochastic bilevel optimization problem in \eqref{eq:stoc_special_case}, regularity constraints on the sequence of objectives are essential \citep{besbes2015non}. These constraints, often expressed as sublinear comparator sequences, quantify the temporal variation in objectives. One key metric in OBO is the \(p\)-th order outer-level function variation, defined as:
\begin{equation}
\label{eq:stochastic_comparator}
\begin{aligned}
V_{p,T} 
&:= \sum_{t=1}^{T-1} \sup_{\outvar \in \Lambda} 
   \Bigl|
   \mathbbE_{(x,y) \sim \mathbb{P}_{t+1}} 
   \bigl[ \mathcal{F}_{t+1}(\outvar, \innerpred^\star_{t+1,\outvar}, x, y) \bigr] \\
&\quad - \mathbbE_{(x,y) \sim \mathbb{P}_t} 
   \bigl[ \mathcal{F}_t(\outvar, \innerpred^\star_{t,\outvar}, x, y) \bigr] 
   \Bigr|^p,
\end{aligned}
\end{equation}
where  \(\mathcal{F}_t(\outvar, \innerpred^\star_{t,\outvar}, x, y) = \pointWoutOBJ(\outvar, \innerpred^\star_{t,\outvar}(x), x, y)\) is the outer objective from \eqref{eq:stoc_special_case}, with \(\mathcal{F}_t = 0\) for \(t < 0\). The metric  \(V_{p,T}\) \citep{lin2023non} measures changes in the outer objective’s expected value. In this work, we focus on first-order outer-level function variation \(V_{1,T} = o(T)\) to derive sublinear regret bounds for our proposed Algorithm~\ref{alg:smooth_fbo_stochastic}.

\section{SmoothFBO: A Generalized FBO Algorithm}

This section introduces the generalized Functional Bilevel Optimization (FBO) algorithm, \textit{SmoothFBO} in Algorithm \ref{alg:online_functional_bo}. Our proposed method extends current algorithms for functional bilevel optimization \citet{petrulionyte2024functional} to non-stationary environments by introducing a stochastic hypergradient estimator constructed via time smoothing, a technique commonly employed in online algorithms \citep{hazan2017efficientregretminimizationnonconvex,lin2023non}. For simplicity, we first introduce a stochastic functional hypergradient oracle that allows us to conveniently analyze the bias and variance properties of our time-smoothed hypergradient estimator in Algorithm~\ref{alg:online_functional_bo}, independent of confounding factors arising from hypergradient estimation.

\begin{definition}[Stochastic Hypergradient Oracle]
\label{def:oracle}
We define a stochastic functional hypergradient oracle $\mathcal{O}(\outvar)$ that returns an i.i.d.\ vector $\outgradhatt(\outvar)$ such that
\begin{align*}
&\mathbb{E}_{\Omega_t}\!\brackets{\outgradhatt(\outvar)} = \outgradt(\outvar), \\
&\text{Var}_{\Omega_t}\!\brackets{\outgradhatt(\outvar)} 
:= \mathbb{E}_{\Omega_t}\!\brackets{\|\outgradhatt(\outvar)\|^2} 
 \nonumber\\&- \bigl\|\mathbb{E}_{\Omega_t}\!\brackets{\outgradhatt(\outvar)}\bigr\|^2 
 \leq \sigma^2_f,
\end{align*}
where $\outgradt(\outvar)$ denotes the true hypergradient given in \eqref{eq:hyper_grad}, and $\Omega_t = \mathbb{P}_t \times \mathbb{Q}_t$.
\end{definition}

At each round $t$, Algorithm~\ref{alg:online_functional_bo} queries the stochastic hypergradient oracle to obtain an estimate $\outgradhatt(\outvar)$ of the true hypergradient $\outgradt(\outvar)$. While single-round stochastic estimates are effective in stationary settings, such estimates are less suitable for non-stationary environments where gradients between rounds may change rapidly. Lemma~\ref{lem:time_smoothed} introduces a time-smoothed stochastic estimate, constructed over a window of length $w$.

\begin{lemma}[Time-Smoothed Hypergradient Estimator]
\label{lem:time_smoothed_oracle}
Under the assumptions of Section~\ref{appx:FBO_assumptions_oracle}, let
\begin{align*}
\outgradtildt(\outvar_t) 
&:= \tfrac{1}{w} \sum_{i=0}^{w-1} \outgradhatti(\outvar_{t-i}),\\
r_{t,w}&:=\tfrac1w\sum_{i=0}^{w-1}
\left(\outgradti(\outvar_{t-i})-\outgradti(\outvar_t)\right).
\end{align*}
With every window summand having time index below one set to zero,
\begin{equation}\label{eq:oracle_tracking_error}
\mathbb E\|\outgradtildt(\outvar_t)-\outgradtw(\outvar_t)\|^2
\leq\frac{2\sigma_f^2}{w}+2\mathbb E\|r_{t,w}\|^2.
\end{equation}
\end{lemma}
Define $D_{T,w}:=\sum_{t=1}^T\mathbb E\|r_{t,w}\|^2$. The proof and exact stored-gradient decomposition are in the Appendix.

\begin{algorithm}[t]
   \caption{\textbf{\emph{SmoothFBO}} (Smooth Functional Bilevel Optimization)}
   \label{alg:online_functional_bo}
\begin{algorithmic}[1]
    \Require Initial outer parameter $\outvar_1$, step size $\alpha > 0$, window size $w \geq 1$, stochastic oracle $\mathcal{O}(\outvar)$
    \For{$t=1,\ldots,T$}
        \State $\outgradhatt(\outvar_t) \gets \mathcal{O}(\outvar_t)$
        \State $\outgradtildt(\outvar_t) \gets \tfrac{1}{w} \sum_{i=0}^{w-1} \outgradhatti(\outvar_{t-i})$
        \State $\outvar_{t+1} \gets \outvar_t - \alpha \outgradtildt(\outvar_t)$
    \EndFor
\end{algorithmic}
\end{algorithm}

\begin{definition}[Bilevel Local Regret]
\label{def:bilevel_local_regret}
Given window length $w \geq 1$ and outer variables $\{\outvar_t\}_{t=1}^T$, define
\begin{align*}
\text{BLR}_{w}(T) := \sum_{t=1}^T \bigl\| \outgradtw(\outvar_t)\bigr\|^2,
\end{align*}
where $\mathcal{F}_{t,w}(\outvar) := \tfrac{1}{w} \sum_{i=0}^{w-1} \mathcal{F}_{t-i}(\outvar)$ with $\mathcal{F}_t = 0$ for $t<1$.
\end{definition}

\begin{theorem}\label{thrm:gd_bilevel_oracle}
Under the assumptions of Section~\ref{appx:FBO_assumptions_oracle}, Algorithm~\ref{alg:online_functional_bo} with step size $\alpha = 1/L$ satisfies
\begin{align*}
\mathbb E[\text{BLR}_{w}(T)]
= \sum_{t=1}^T \mathbb E
   \!\left[\|\outgradtw(\outvar_t)\|^2\right]
\;\leq\; \nonumber\\ 2L \!\left( \tfrac{2TQ}{w} + V_{1,T} 
   + \tfrac{T \sigma^2_f}{Lw}+\tfrac{D_{T,w}}{L} \right).
\end{align*}
\end{theorem}

\begin{corollary}
For window size $w=w_T$ such that $w_T\to\infty$ and $w_T=o(T)$, if $D_{T,w}=o(T)$, the expected regret is sublinear. In particular,
\begin{align*}
\sum_{t=1}^T \mathbbE \!\left[ 
   \|\nabla \mathcal{F}_{t,w}(\outvar_t)\|^2 \right] 
   \;\leq\; O \!\left( \tfrac{TQ}{w} 
   + V_{1,T} + \tfrac{T \sigma^2_f}{w}+D_{T,w} \right).
\end{align*}
\end{corollary}

\section{SmoothFBO with Hypergradient Estimation}
This section presents Algorithm 2, an extension of \textbf{SmoothFBO} (Algorithm 1), which replaces the hypergradient oracle with a time-smoothed stochastic functional hypergradient estimator for the outer objective \(\mathcal{F}_t(\outvar) = \mathbb{E}_{(x,y) \sim \mathbb{P}_t} \left[ \ell_{out}(\outvar, \innerpred^\star_{t,\outvar}(x), x, y) \right]\). Building on the original time-smoothed estimator of Lemma \eqref{lem:time_smoothed}, this approach averages stochastic functional hypergradient estimates from the following algorithm (Algorithm \ref{alg:func_grad}) over a window of size \( w \), achieving variance reduction for approximate stochastic gradients similar to the oracle setting, as established in Theorem~\ref{thrm:alg1_rate}. Algorithm \ref{alg:func_grad} is for functional hypergradient estimation from \cref{eq:hyper_grad}, the subroutines for finding the adjoint function (\texttt{AdjointOpt}) and the inner prediction function (\texttt{InnerOpt}) can be found in \cref{appx:functional_hypergrad_algo}. \begin{algorithm}[htb] \caption{\textbf{FuncGrad}(\(\outvar, \innerpred, \adjoint, \mathcal{D}\))} \label{alg:func_grad} \begin{algorithmic} \Require current outer, inner, and adjoint models $\outvar$, $\innerpred$, $\adjoint$, dataset $\mathcal{D} = (\inDataset,\outDataset)$ \State \textcolor{darkred}{ \#\ \textit{Inner-level optimization}} \State $\hat{\innerpred}_{\boldsymbol{\lambda}} \leftarrow $ {\bf\texttt{InnerOpt}}($\outvar,\innerpred,\inDataset$) \State \textcolor{darkred}{ \#\ \textit{Adjoint optimization}} \State $\hat{\adjoint}_{\outvar} \leftarrow $ {\bf \texttt{AdjointOpt}($\outvar,\adjoint,\hat{\innerpred}_{\outvar},\mathcal{D}$)} \State \textcolor{darkred}{ \#\ \textit{Hypergradient estimation}} \State Sample a mini-batch $\mathcal{B}=(\outBatch,\inBatch)$ from $\mathcal{D}=(\outDataset,\inDataset)$ \State $g_{Exp}\leftarrow \partial_{\outvar}\empoutOBJ(\outvar,\hat{\innerpred}_{\outvar},\outBatch)$ \State $g_{Imp}\leftarrow \frac{1}{\verts{\inBatch}} \sum_{(x,y)\in \inBatch}\partial_{\outvar,v} \pointWinOBJ(\outvar, \hat{\innerpred}_{\outvar}(x),x,y)\ \hat{\adjoint}_{\outvar}(x)$ \State \Return $g_{Exp}+g_{Imp},\hat{\innerpred}_{\outvar},\hat{\adjoint}_{\outvar}$ \end{algorithmic} \end{algorithm} 

\begin{lemma}[Time-Smoothed Hypergradient Estimator] \label{lem:time_smoothed} Under the estimation-error assumptions in Section~\ref{appx:SmoothFBO_assumptions}, the error of the stored estimator $\outgradtildt(\outvar_t)$ relative to $\outgradtw(\outvar_t)$ decomposes into averaged centered noise, averaged hypergradient bias, and the tracking term $r_{t,w}$. The exact decomposition is in the Appendix. \end{lemma}

\begin{algorithm}[htb] \caption{\textbf{SmoothFBO} (Smooth Functional Bilevel Optimization)} \label{alg:smooth_fbo_stochastic} \begin{algorithmic} \Require Step  \(\alpha > 0\), window size \( w \geq 1\), data distribution \(\mathbb{P}_t\),  hypergradient estimator \textbf{FuncGrad}(\(\outvar, \innerpred, \adjoint,\mathbb{P}_t\)), initial  $\outvar_1$ and models $\innerpred_{\outvar_{1}}, \adjoint_{\outvar_{1}}$ \For{$t=1,\ldots,T$} \State \textcolor{darkred}{\#\ Query stochastic hypergradient estimate} \State \(\outgradhatt(\outvar_t), \innerpred_{\outvar_{t+1}}, \adjoint_{\outvar_{t+1}} \gets  \textbf{FuncGrad}(\outvar_t, \ldots)\) \State \textcolor{darkred}{\#\ Compute time-smoothed stochastic estimator} \State \(\outgradtildt(\outvar_t) \gets \frac{1}{w} \sum_{i=0}^{w-1} \outgradhatti(\outvar_{t-i})\) \State \textcolor{darkred}{\#\ Update outer parameter} \State \(\outvar_{t+1} \gets \outvar_t - \alpha \outgradtildt(\outvar_t)\) \EndFor \end{algorithmic} \end{algorithm}

\begin{lemma}[Expected Squared Error of Time-Smoothed Hypergradient Estimator] \label{lem:time_smoothed_error} Let $\outgradtildt(\outvar_t)$ denote the time-smoothed hypergradient estimator defined in \ref{lem:time_smoothed}, where $\outgradhatti(\outvar_{t-i})$ is the stochastic hypergradient estimate from Algorithm~\ref{alg:func_grad} at time $t-i$. The expected error is bounded by \begin{align*} &\mathbb E \left[ \sum_{t=1}^T \Verts{ \outgradtildt(\outvar_t) - \outgradtw(\outvar_t) }^2 \right] \nonumber\\&\leq C_1 \frac{T \sigma^2_{\mathcal{F}}}{w} + C_2 \sum_{t=1}^T \epsilon_{\text{in},t} + C_3 \sum_{t=1}^T \epsilon_{\text{adj},t}+3D_{T,w}, \end{align*} where $C_1$, $C_2$, and $C_3$ are constants defined in the Appendix, $\epsilon_{\text{in},t}$ and $\epsilon_{\text{adj},t}$ represent the inner and adjoint approximation errors, respectively, and $w$ is the window size. A proof is in the Appendix. \end{lemma} This bound on the expected squared error of $\outgradtildt(\outvar_t)$ ensures that the time-smoothed hypergradient estimator remains accurate under sublinear approximation errors $\epsilon_{\text{in},t}$ and $\epsilon_{\text{adj},t}$, enabling us to leverage these properties in the subsequent analysis to establish sublinear $\textBLR_w(T)$ in the convergence of Algorithm~\ref{alg:smooth_fbo_stochastic}. Next, we introduce an additional assumption required for our regret analysis. \begin{enumerate} \item \textbf{Approximate Optimality with Sublinear Errors}: The inner optimization and adjoint problems have sublinear approximation errors $\epsilon_{\text{in},t}$ and $\epsilon_{\text{adj},t}$ across time, satisfying: \[ \sum_{t=1}^T \epsilon_{\text{in},t}^2 = o(T) \quad \text{and} \quad \sum_{t=1}^T \epsilon_{\text{adj},t}^2 = o(T). \] \end{enumerate} 

\begin{theorem}\label{thrm:alg1_rate} Under the assumptions of Section~\ref{appx:SmoothFBO_assumptions}, the bilevel local regret of Algorithm~\ref{alg:smooth_fbo_stochastic}, using the time-smoothed hypergradient estimator $\outgradtildt(\outvar_t)$, achieves an upper bound with step size $\alpha = \frac{4}{5L}$: \begin{align} &\mathbb E[\textBLR_w(T)] = \sum_{t=1}^T \mathbb E \brackets{ \Verts{ \outgradtw(\outvar_t) }^2 } \notag \\ &\leq C_4 \parens{ \frac{2TQ}{w} + V_{1,T} } + C_5\mathbb E  \Gamma_{T,w}  \notag \\ &\leq C_4 \parens{ \frac{2TQ}{w} + V_{1,T} } + C_5 \tilde{\Gamma}_{T,w} , \end{align} where for shorthand we denote $\Gamma_{T,w}:=\sum_{t=1}^T \Verts{ \outgradtildt(\outvar_t) - \outgradtw(\outvar_t) }^2$, with an upper bound $\tilde{\Gamma}_{T,w}:=C_1 \frac{T\sigma^2_{\mathcal{F}}}{w} + C_2 \sum_{t=1}^T \epsilon_{\text{in},t} + C_3 \sum_{t=1}^T \epsilon_{\text{adj},t}+3D_{T,w}$, $L$ is the Lipschitz constant of $\outgradt$, $\sigma^2_{\mathcal{F}}$ is the variance bound of the hypergradient estimates, $Q$ bounds the outer objective, $V_{1,T} = o(T)$ quantifies the variation in the comparator sequence, and $C_1$, $C_2$, $C_3$ are constants from Lemma~\ref{lem:time_smoothed_error} associated with the approximation errors $\epsilon_{\text{in},t}$ and $\epsilon_{\text{adj},t}$. For window size $w=w_T$ with $w_T\to\infty$ and $w_T=o(T)$, the regret $\textBLR_w(T)$ of Algorithm~\ref{alg:smooth_fbo_stochastic} is sublinear when $\sum_{t=1}^T \epsilon_{\text{in},t}^2 = o(T)$, $\sum_{t=1}^T \epsilon_{\text{adj},t}^2 = o(T)$, and $D_{T,w}=o(T)$. The proof with constants $C_4,C_5$ is provided in the Appendix. \end{theorem} Having established an upper bound on the bilevel local regret $\textBLR_w(T)$ for Algorithm~\ref{alg:smooth_fbo_stochastic}, which generalizes the oracle setting by incorporating the hypergradient estimation errors of $\outgradtildt(\outvar_t)$, the following corollary describes the resulting variance--tracking trade-off.

\begin{corollary} \label{cor:var_red} Increasing the window parameter $w$ in Algorithm~\ref{alg:smooth_fbo_stochastic} reduces the displayed variance contribution but may increase the tracking error, as given by: \begin{align} \sum_{t=1}^T \mathbb E \brackets{ \Verts{ \outgradtw(\outvar_t) }^2 } \nonumber\\\leq \bigO \parens{ \frac{TQ}{w} + V_{1,T} + \frac{T \sigma^2_{\mathcal{F}}}{w} + \sum_{t=1}^T \epsilon_{\text{in},t} + \sum_{t=1}^T \epsilon_{\text{adj},t}+D_{T,w} }, \end{align} where a larger $w$ diminishes the variance term $\frac{T \sigma^2_{\mathcal{F}}}{w}$, while sublinear regret additionally requires $D_{T,w}=o(T)$. \end{corollary}

\begin{lemma}[Reduction of Rates with Linear Inner Predictor] \label{lem:linear_inner} Consider the case where the inner predictor is linear, \( \innerpred^\star_{t,\outvar}(x) = \Phi(x)\,\theta^\star_{t,\outvar}, \) where $\theta^\star_{t,\outvar}$ is the optimal parameter obtained from the inner optimization problem and $\Phi(x)$ is a linear mapping of $x$. In this setting, the online functional bilevel optimization problem \eqref{def:FBO} reduces to the parametric special case, analyzed within \citet{bohne2024online}. Under the assumptions of Section~\ref{appx:SmoothFBO_assumptions} and the historical-point premise stated in the Appendix, the bilevel local regret of Algorithm~\ref{alg:smooth_fbo_stochastic} then satisfies \begin{align} &\sum_{t=1}^T \mathbb E \brackets{ \Verts{ \outgradtw(\outvar_t) }^2 } \nonumber\\&\leq \bigO \parens{ \frac{TQ}{w} + V_{1,T} + \frac{T \sigma^2_{\mathcal{F}}}{w} +H_{2,T}+D_{T,w}} \end{align} where the comparator sequence of $H_{2,T} $ is the second-order path variation from the parametric OBO setting defined as $H_{2,T}:=\sum_{t=2}^T\sup_{\boldsymbol{\lambda}\in \Lambda}\left\|\theta^*_{t-1,\boldsymbol{\lambda}}-\theta^*_{t,\boldsymbol{\lambda}}\right\|^2$. For window size $w=w_T$ with $w_T\to\infty$ and $w_T=o(T)$, the regret $\textBLR_w(T)$ of Algorithm~\ref{alg:smooth_fbo_stochastic} is sublinear under the standard conditions that comparator sequences satisfy regularity constraints $V_{1,T}=o(T)$, $H_{2,T}=o(T)$, and $D_{T,w}=o(T)$, see \citet{tarzanagh2024online, lin2023non}. A proof for this lemma is found in the Appendix. \end{lemma}

\section{Experiments}\label{sec:experiments}

\subsection{Importance–weight tuning for non-stationary regression}
\label{sec:synth}

We begin with a controlled regression benchmark that instantiates the non-stationary functional setup in \eqref{eq:stoc_special_case}. 
The ground-truth data-generating process is a single-neuron sigmoid network:
\begin{equation}
    f_{\text{true}}(x_t) = \sigma(W_t^\top x + b_t),
\end{equation}
where the underlying parameters $(W_t, b_t)$ evolve nonstationarily over time. 
Figure~\ref{fig:drift} (Fig.~2) summarizes the temporal drift of these parameters in the data-generating process (DGP) via a sinusoidal drift $\beta\sin(\omega t)$.
Inputs are drawn i.i.d.\ as $\mathbf{X}_t\sim\mathcal{N}(0,\mathbf{I})$, and observed targets are
\begin{equation}
    \mathbf{Y}_t = f_{\text{true}}(\mathbf{X}_t)  + \boldsymbol{\zeta}_t,
\end{equation}
with Gaussian noise 
$\boldsymbol{\zeta}_t \sim \mathcal{N}(0,\sigma^2 \mathbf{I})$. The inner predictor is a three-layer MLP $h\in\Hcal$ with GeLU activations. 
The outer variable is a nonnegative scalar importance weight $\lambda \in \Real_{\ge 0}$ that controls the emphasis on recent versus older minibatches. 
Given a window $W_t=\{t-w,\dots,t-1\}$ of length $w\ge 1$, we solve the bilevel problem
\begin{align}
&\min_{\lambda\in\Lambda}\quad
\outOBJ(\lambda,\innerpred^\star_{t,\lambda})
:= \tfrac{1}{B}\!\sum_{i=1}^B \big\|Y_{t,i}-\innerpred^\star_{t,\lambda}(X_i)\big\|_2^2, \label{eq:synth_outer}\\
&\text{s.t.}\quad
\innerpred^\star_{t,\lambda} \approx \argmin_{\innerpred\in\Hcal}
\sum_{s\in W_t} \lambda_{t,s}\cdot
\tfrac{1}{B}\!\sum_{i=1}^B \big\|Y_{s,i}-\innerpred(X_i)\big\|_2^2, \label{eq:synth_inner}
\end{align}
where nonnegative weights $\lambda_{t,s}$ are projected as $\lambda\leftarrow\max(\lambda,0)$ after each update.
The outer loss \eqref{eq:synth_outer} uses a holdout minibatch at time $t$, while the inner loss \eqref{eq:synth_inner} aggregates minibatches from the sliding window.

\paragraph{Baselines.} 
We compare the following hypergradient estimators:
\begin{enumerate}[leftmargin=*]
    \item \textbf{Parametric} (unrolling truncated backprop through optimization),
    \item \textbf{FBO} (functional bilevel optimization) without window-smoothing of \citet{petrulionyte2024functional}
    \item \textbf{SmoothFBO} (ours), which averages stochastic functional hypergradients over a window of length $w$ before the outer update.
\end{enumerate}
Further comparisons to offline and online parametric baselines such as \emph{Approximate Implicit Differentiation (AID)} are deferred to the appendix.

\begin{figure}[t]
    \centering
    \includegraphics[width=\columnwidth]{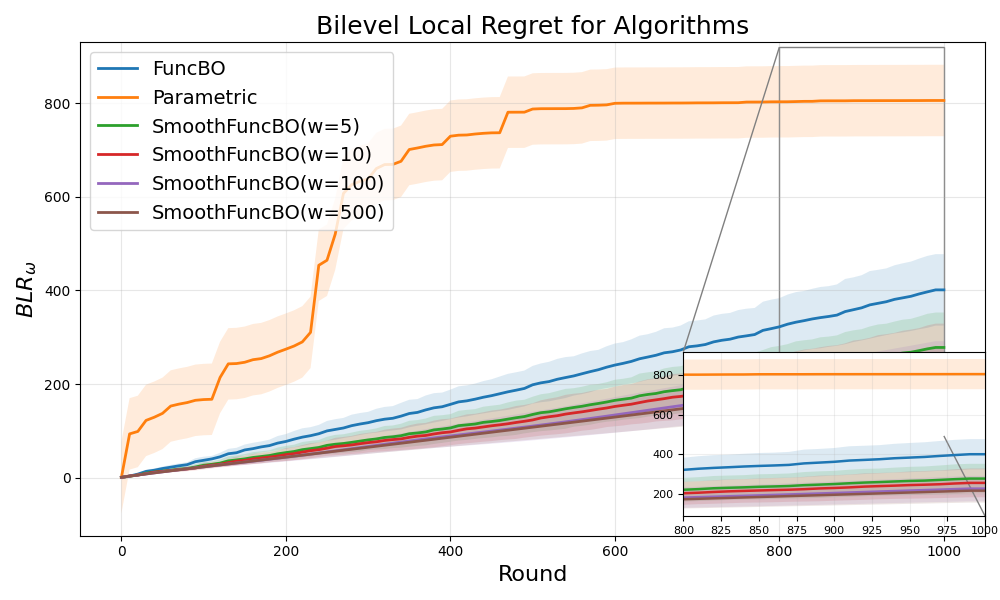}
    \vspace{-6pt}
    \caption{Cumulative stored-gradient BLR proxy vs.\ rounds (Fig.~1).
    SmoothFBO exhibits a sublinear-looking empirical trend.
    The zoomed-in component highlights this trend, while increasing the window $w$ ($5\!\to\!500$) further reduces the proxy.}
    \label{fig:synth}
\end{figure}

\begin{figure}[t]
    \centering
    \includegraphics[width=\columnwidth]{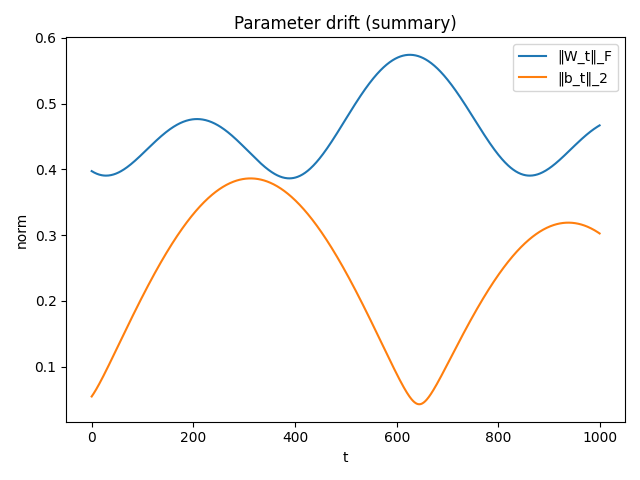}
    \vspace{-6pt}
    \caption{Parameter drift in the underlying DGP (Fig.~2). 
    The weights $(W_t, b_t)$ evolve nonstationarily, driving the outer-loop adaptation challenge.}
    \label{fig:drift}
\end{figure}

\paragraph{Results.}
Figure~\ref{fig:synth} (Fig.~1) reports the cumulative stored-gradient BLR proxy, $\sum_{t}\|\outgradtildt(\outvar_t)\|^2$, versus rounds; its relation to Definition~\ref{def:bilevel_local_regret} is given in the Appendix.
Empirically, \textbf{SmoothFBO} exhibits a \emph{sublinear-looking proxy trend}, as highlighted in the zoomed-in panel.
Moreover, increasing the window size $w$ (from $5$ to $500$) further reduces the proxy.
In contrast, FBO and parametric methods incur substantially larger proxy values.

Figure~\ref{fig:drift} (Fig.~2) highlights the temporal evolution of $(W_t, b_t)$ in the data-generating process, which induces the nonstationary regression challenge. 
Further analysis in the Appendix analyzes the loss and gradient norms across considered algorithms as well additional ablation studies on the experiment design. 

\subsection{Non-Stationary Model-based Reinforcement Learning}
\label{sec:experiments-rl}

Non-stationary reinforcement learning environments present significant challenges as the underlying system dynamics evolve over time, rendering traditional RL approaches ineffective due to their stationarity assumptions. Our experiments demonstrate that the non-stationary functional bilevel optimization framework effectively captures these time-varying dynamics.

\paragraph{Problem formulation.} In model-based RL with non-stationary environments, the time-varying Markov Decision Process (MDP) is approximated by a probabilistic model $q_{\outvar,t}$ with parameters $\outvar$. The model predicts the next state $s_{\outvar,t}(x)$ and reward $r_{\outvar,t}(x)$, given a pair $x:=(s,a)$ where $s$ is the current environment state and $a$ is the agent's action. 

A second model approximates the action-value function $\innerpred_t(x)$ that computes the expected cumulative reward given the current state-action pair at time $t$. Traditionally, the action-value function is learned using the current MDP model, while the MDP model is learned independently using Maximum Likelihood Estimation (MLE) \citep{sutton1991dyna}. However, in recent work, \citet{Nikishin:2022} showed that casting model-based RL as a bilevel problem can result in better performance and tolerance to model-misspecification (see \ref{appx:experiment}).

In our online bilevel formulation, the inner-level problem at time $t$ involves learning the optimal action-value function $\innerpred_{t,\outvar}^{\star}$ with the current MDP model $q_{\outvar,t}$ by minimizing the Bellman error. The inner-level objective can be expressed as an expectation of a point-wise loss $f$ with samples $(x,r',s')\sim \mathbb{Q}_t$, derived from the agent-environment interaction at time $t$:
\begin{align}\label{eq:RL_bielvel_inner}
    \innerpred_{t,\outvar}^{\star} = \arg\min_{\innerpred \in \Hcal} \mathbb{E}_{\mathbb{Q}_t}\brackets{f(\innerpred(x),r_{\outvar,t}(x),s_{\outvar,t}(x))}.
\end{align}

\vspace{-0.1cm}
Here, the future state and reward $(r',s')$ are replaced by the time-varying MDP model predictions $r_{\outvar,t}(x)$ and $s_{\outvar,t}(x)$. In practice, samples from $\mathbb{Q}_t$ are obtained using a replay buffer that adapts to the changing environment dynamics. The buffer accumulates data by interacting with the environment at time $t$. The non-stationarity in our environment is modeled by shifting the pole angle reward zones with time, which fundamentally alters the system's dynamics at each time step $t$. This creates a sequence of time-varying MDPs that the agent must continuously adapt to, the exact setup is further detailed in \ref{appx:experiment}.

The point-wise loss function $f$ represents the error between the action-value function prediction and the expected cumulative reward given the current state-action pair: 
\begin{align*}
f(v, r',s') := \frac{1}{2}\Verts{ v - r' - \gamma \log\sum_{a'}e^{\bar{h}_t(s', a')}}^2,
\end{align*}
with $\bar{h}_t$ a lagged version of $\innerpred_t$ and $\gamma$ a discount factor.

The time-varying MDP model is learned implicitly using the optimal function $\innerpred_{t,\outvar}^{\star}$, by minimizing the Bellman error w.r.t.~$\outvar$ at each time step $t$: 
\begin{align}\label{eq:RL_bielvel_outer}
     \min_{\outvar\in\Lambda}\ &\mathbb{E}\brackets{f(\innerpred_{t,\outvar}^{\star}(x), r',s')}.
\end{align}

\vspace{-0.1cm}
Equations \ref{eq:RL_bielvel_outer} and \ref{eq:RL_bielvel_inner} define a non-stationary bilevel problem as in the general framework of equation \ref{eq:stoc_special_case}, where at each time step $t$, we have data distribution $\mathbb{Q}_t = \mathbb{P}_t$, $y=(r',s')$, and the point-wise losses $\pointWinOBJ$ and $\pointWoutOBJ$ are given by: $\pointWinOBJ\parens{\outvar,v,x,y}= f\parens{v, r_{\outvar,t}(x),s_{\outvar,t}(x)}$ and $\pointWoutOBJ\parens{\outvar, v,x,y}=f\parens{v,r',s'}$. Therefore, we can directly apply our SmoothFBO \cref{alg:smooth_fbo_stochastic} to learn both the time-varying MDP model $q_{\outvar,t}$ and the optimal action-value function $\innerpred_{t,\outvar}^{\star}$.

\paragraph{Experimental details.} 
We evaluate the proposed \textit{SmoothFBO} algorithm against three baselines and their time-smoothed variants: 
\begin{enumerate}[leftmargin=*]
    \item \textbf{Maximum Likelihood Estimation (MLE):} The standard approach that updates the world model by direct likelihood maximization \citep{sutton1991dyna}.
    \item \textbf{Optimal Model Design (OMD):} A parametric bilevel method for RL following implicit differentiation \citep{Nikishin:2022}.
    \item \textbf{Iterative Differentiation (ITD):} Differentiating through the inner optimizer \citep{Lorraine2019OptimizingMO}.
    \item \textbf{Functional Bilevel Optimization (FBO):} The functional approach of \citet{petrulionyte2024functional} without temporal smoothing.
    \item \textbf{SmoothFBO (ours):} An extension of FBO with time-smoothed hypergradient estimation.
\end{enumerate}

\begin{figure}[t]
    \centering
    \includegraphics[width=0.7\columnwidth]{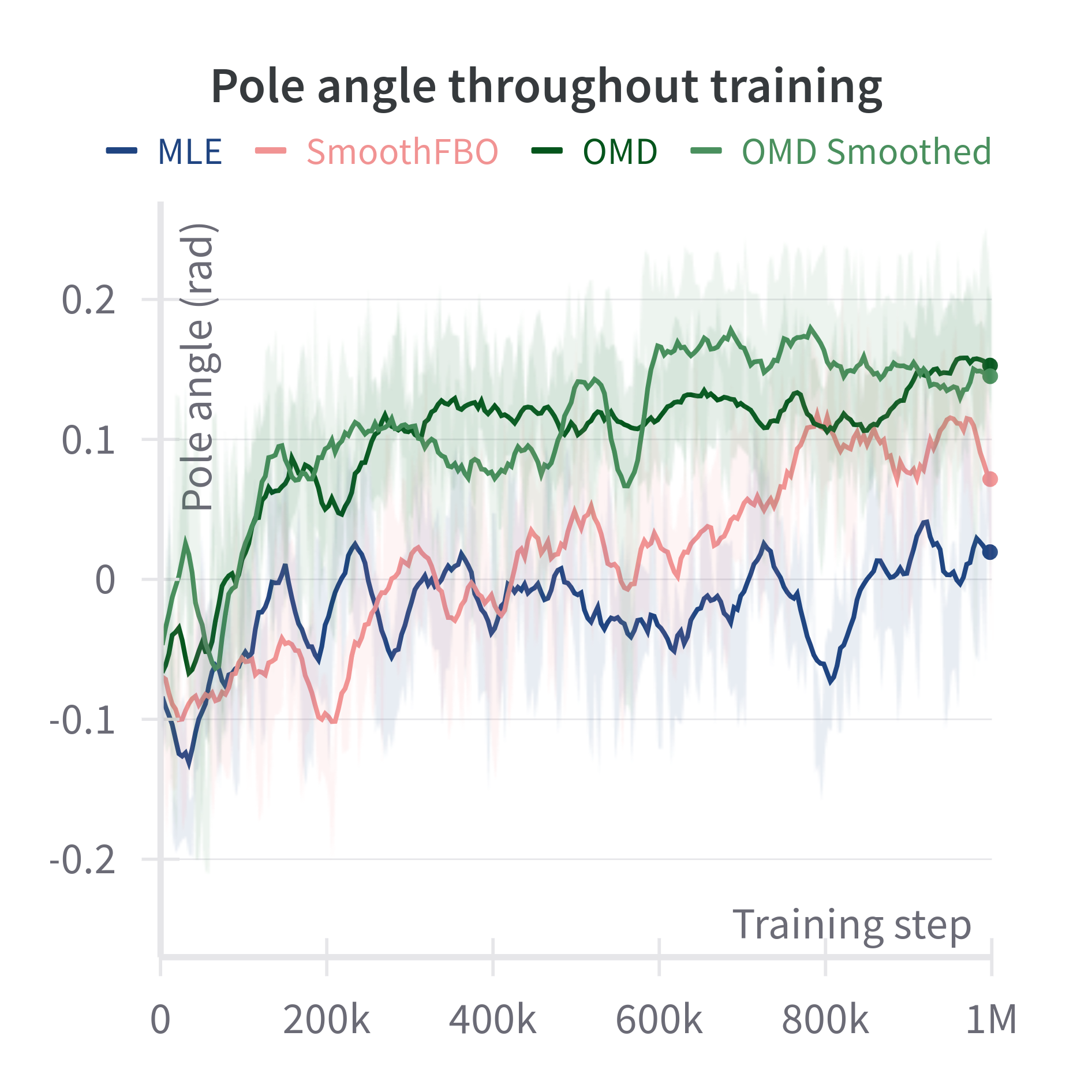}
    \vspace{-6pt}
    \caption{The changing pole angle throughout training. 
    In the non-stationary CartPole experiment, the target pole angle shifts gradually over training steps, forcing the agent to adapt its learned dynamics and policy.}
    \label{fig:non_s_reward}
\end{figure}

To create a challenging non-stationary testbed, we implemented a modified CartPole environment \citep{brockman2016openai} where the reward structure drifts gradually over time. 
In the stationary environment, the agent is rewarded for maintaining the pole angle within a fixed optimal interval. 
In our non-stationary variant, this interval interpolates smoothly between two different regions, requiring the agent to adapt both its world model and policy continually. 
Figure~\ref{fig:non_s_reward} shows the evolving pole angle target region during training. 
This design makes adaptation essential: the change is large enough to invalidate static models, but gradual enough that tracking is feasible. 

Our evaluation protocol consists of two phases: (1) hyperparameter tuning via grid search in the stationary environment, and (2) comprehensive evaluation of the best configurations across multiple random seeds in both stationary and non-stationary settings. 
All results are averaged across $5$–$10$ seeds, with shaded regions in the plots indicating variability. 
Implementation and hardware details are deferred to \S\ref{appx:experiment}.

\vspace{-0.2cm}
\subsection{Results and Analysis}

\begin{figure}[t]
    \centering
    \includegraphics[width=0.49\columnwidth]{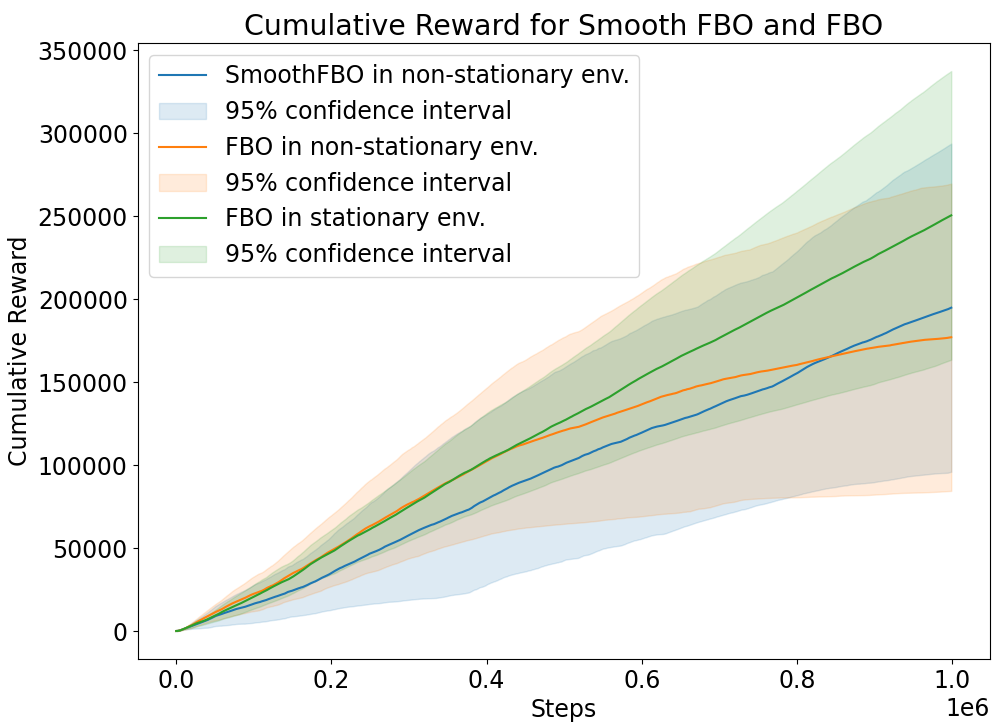}
    \includegraphics[width=0.49\columnwidth]{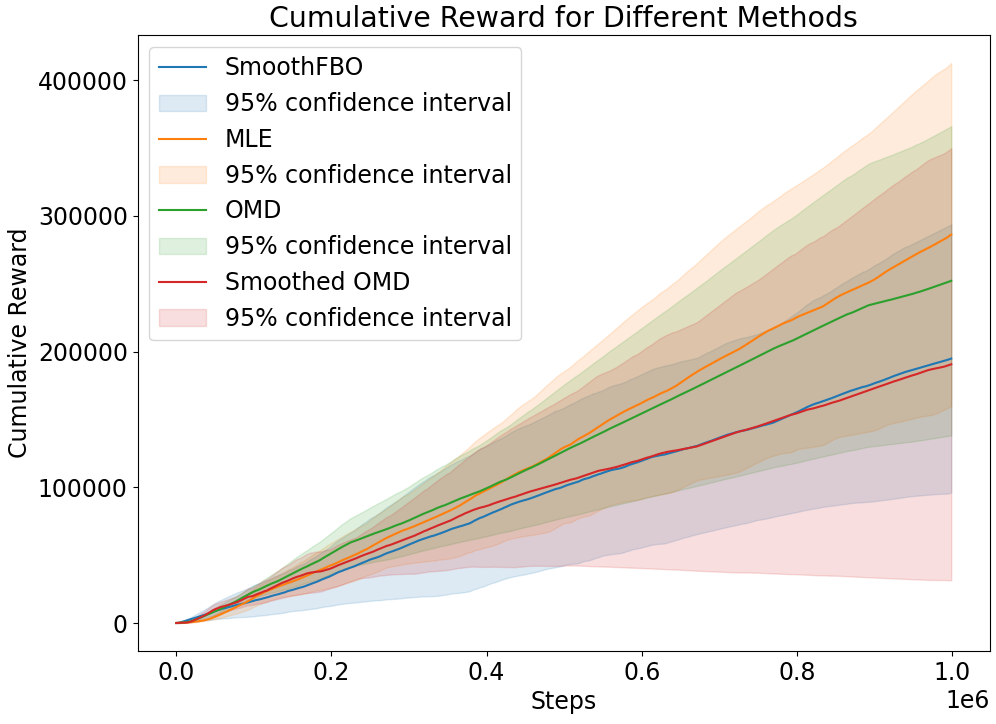}
    \vspace{-6pt}
    \caption{Cumulative reward for the non-stationary CartPole evaluation environment over 1 million environment steps. Each curve represents the mean cumulative episode reward across 10 random seeds, with shaded regions indicating 95\% confidence intervals. \textbf{Left}: The FBO method in stationary and non-stationary environments compared to \textit{SmoothFBO}. \textbf{Right}: Comparison with baseline methods where \textit{SmoothFBO} matches their performance in adapting to the non-stationary dynamics with an averaging window of 100.}
    \label{fig:cartpole}
 \end{figure}

\paragraph{Evaluation.} 
After tuning in the stationary setting, we evaluate each algorithm’s best configuration in both stationary and non-stationary environments. 
Performance is measured by cumulative episode reward. 
Figure~\ref{fig:cartpole} summarizes the results. 

On the left, we observe that in the stationary case both FBO and \textit{SmoothFBO} perform competitively; however, when the reward structure drifts, \textit{SmoothFBO} is significantly more robust and maintains higher cumulative rewards. 
The right panel compares all baselines: \textit{SmoothFBO} matches or exceeds their performance, highlighting its advantage in adapting to non-stationary dynamics.

We additionally compare against parametric unrolling/ITD \citep{iter_bilevel} with and without time-smoothing, and report the final episode reward on the evaluation environment in Table~\ref{tab:rl_additional_baselines}. In our non-stationary CartPole setup, while ITD can occasionally achieve partial convergence (mean final reward $56.87$, max $336.05$), it remains much less stable across seeds compared to \textit{SmoothFBO}. While performing a grid search on ITD, we obtained notoriously unstable hypergradients with Adam, the optimizer used for all other methods, and had to switch to SGD. Even then, ITD remains less stable than the other approaches.

These findings emphasize two key points: (i) temporal smoothing in \textit{SmoothFBO} improves stability under drift, and (ii) robustness to non-stationarity does not compromise efficiency in the stationary setting. 

\begin{table}[t]
\centering
\label{tab:rl_additional_baselines}
\begin{tabular}{lccc}
\toprule
Method & Max & Mean & Min \\
\midrule
Smoothed ITD & 23.40 & 9.56 & 4.20 \\
ITD & 336.05 & 56.87 & 4.20 \\
Smoothed OMD & 147.75 & 31.55 & 5.00 \\
\textit{SmoothFBO} & \textbf{500.00} & \textbf{232.19} & \textbf{5.05} \\
\bottomrule
\end{tabular}
\caption{Final episode reward on the evaluation environment in the non-stationary CartPole experiment, over 10 seeds, for additional parametric bilevel baselines and representative implicit baselines. ITD/unrolling exhibits high variability and lower reliability than \textit{SmoothFBO}.}
\end{table}

\section{Conclusion}

This work presents a non-stationary functional bilevel optimization framework \textit{SmoothFBO}. This method extends functional bilevel optimization to non-stationary environments through time-smoothing techniques that reduce variance in the outer loop. This enables more stable convergence with proven sublinear regret bounds. Despite these advances, achieving optimal performance in dynamic reinforcement learning environments remains challenging, as reinforcement learning models remain sensitive to initialization and prone to catastrophic forgetting. Nevertheless, experimental results demonstrate SmoothFBO's practical effectiveness in adapting to changing dynamics where standard FBO methods struggle. Our work opens the door to future research in new variance reduction strategies for bilevel algorithms in non-stationary environments.

\section*{Acknowledgments}
\vspace{-0.2cm}
This work was supported by the ERC grant number 101087696 (APHELAIA project) and by ANR 3IA MIAI@Grenoble Alpes (ANR-19-P3IA-0003) and the ANR project BONSAI (grant ANR-23-CE23-0012-01).

\bibliography{biblio.bib}

\newpage

\appendix

\onecolumn
\section{Convergence Analysis}
\subsection{Assumptions for Algorithm \ref{alg:online_functional_bo}}\label{appx:FBO_assumptions_oracle}
For the convergence analysis of Algorithm~\ref{alg:online_functional_bo}, assume:
\begin{enumerate}
    \item \textbf{Differentiability and iterates in the domain}: For $t=1,\ldots,T$, $\mathcal F_t$ is continuously differentiable on $\Lambda$, its gradient is $L$-Lipschitz, and the iterates $\outvar_1,\ldots,\outvar_{T+1}$ remain in $\Lambda$.
    \item \textbf{Oracle noise}: Let $\{\mathcal A_t\}_{t\geq0}$ be the full algorithmic filtration, where $\mathcal A_{t-1}$ contains all randomness revealed before the $t$th oracle output and $\outvar_t$ is $\mathcal A_{t-1}$-measurable. Then $\xi_t=\outgradhatt(\outvar_t)-\outgradt(\outvar_t)$ is $\mathcal A_t$-measurable and satisfies
    \[
    \mathbb E[\xi_t\mid\mathcal A_{t-1}]=0,\qquad
    \mathbb E[\|\xi_t\|^2\mid\mathcal A_{t-1}]\leq\sigma_f^2.
    \]
    \item \textbf{Bounded objective and zero padding}: $|\mathcal F_t(\outvar)|\leq Q$ for every $\outvar\in\Lambda$ and $t\geq1$. For $s<1$, set $\mathcal F_s\equiv0$, $\xi_s=0$, and define the entire stored-gradient or tracking summand indexed by $s$ to be zero, so no iterate $\outvar_s$ is required.
    \item \textbf{Gradual non-stationarity}: The population-objective variation $V_{1,T}$ in \eqref{eq:stochastic_comparator} is finite. Sublinear regret uses $V_{1,T}=o(T)$.
    \item \textbf{Window and tracking}: $1\leq w\leq T$. The bound holds for any $D_{T,w}$; sublinear regret additionally uses $D_{T,w}=o(T)$.
\end{enumerate}

\subsection{Convergence Analysis for Algorithm \ref{alg:online_functional_bo}}

\begin{proof}[Proof of Lemma~\ref{lem:time_smoothed_oracle}]
Adding and subtracting the historical exact hypergradients gives the pathwise identity
\begin{align*}
\outgradtildt(\outvar_t)-\outgradtw(\outvar_t)
&=\frac1w\sum_{i=0}^{w-1}
  \bigl(\outgradhatti(\outvar_{t-i})-\outgradti(\outvar_{t-i})\bigr)\\
&\quad+\frac1w\sum_{i=0}^{w-1}
  \bigl(\outgradti(\outvar_{t-i})-\outgradti(\outvar_t)\bigr)\\
&=\frac1w\sum_{i=0}^{w-1}\xi_{t-i}+r_{t,w}.
\end{align*}
For $s<u$, $\xi_s$ is $\mathcal A_{u-1}$-measurable, so the martingale-difference property gives
\[
\mathbb E\langle\xi_s,\xi_u\rangle
=\mathbb E\!\left[
 \left\langle\xi_s,\mathbb E[\xi_u\mid\mathcal A_{u-1}]\right\rangle
 \right]=0.
\]
Therefore, using the zero-padding convention,
\[
\mathbb E\left\|\frac1w\sum_{i=0}^{w-1}\xi_{t-i}\right\|^2
=\frac1{w^2}\sum_{i=0}^{w-1}\mathbb E\|\xi_{t-i}\|^2
\leq\frac{\sigma_f^2}{w}.
\]
Finally, $\|a+b\|^2\leq2\|a\|^2+2\|b\|^2$ proves
\eqref{eq:oracle_tracking_error}.
\end{proof}

By $L$-Lipschitzness of the hypergradients and Jensen's inequality, the tracking budget also satisfies
\begin{equation}\label{eq:tracking_movement_bound}
D_{T,w}\leq\frac{L^2}{w}\sum_{t=1}^T\sum_{i=0}^{w-1}
\mathbb E\|\outvar_{t-i}-\outvar_t\|^2,
\end{equation}
with nonpositive-time summands interpreted as zero. Thus $D_{T,w}$ measures the price of using gradients stored at stale iterates.

We next give the moving-window telescoping lemma used by both convergence theorems.
\begin{lemma}\label{lem:stochastic_bounded_time_varying}
Suppose $|\mathcal F_t(\outvar)|\leq Q$ on $\Lambda$, set $\mathcal F_s\equiv0$ for $s<1$, and let $1\leq w\leq T$. For any sequence $\{\outvar_t\}_{t=1}^{T+1}\subseteq\Lambda$,
\begin{align}
\sum_{t=1}^T\left(
\mathcal F_{t,w}(\outvar_t)-\mathcal F_{t,w}(\outvar_{t+1})
\right)
\leq\frac{2TQ}{w}+V_{1,T}.
\end{align}
\end{lemma}

\begin{proof}
An index shift gives
\begin{align*}
S&:=\sum_{t=1}^T\left(
\mathcal F_{t,w}(\outvar_t)-\mathcal F_{t,w}(\outvar_{t+1})
\right)\\
&=\mathcal F_{1,w}(\outvar_1)-\mathcal F_{T,w}(\outvar_{T+1})
+\sum_{t=2}^T
\left(\mathcal F_{t,w}(\outvar_t)-\mathcal F_{t-1,w}(\outvar_t)\right)\\
&=\mathcal F_{1,w}(\outvar_1)-\mathcal F_{T,w}(\outvar_{T+1})
+\frac1w\sum_{t=2}^T
\left(\mathcal F_t(\outvar_t)-\mathcal F_{t-w}(\outvar_t)\right).
\end{align*}
The two boundary terms together with the startup terms $2\leq t\leq w$
are at most
\[
\frac Qw+Q+\frac{(w-1)Q}{w}=2Q.
\]
For the remaining terms, set
$\delta_j:=\sup_{\outvar\in\Lambda}
|\mathcal F_{j+1}(\outvar)-\mathcal F_j(\outvar)|$. Then
\[
\frac1w\sum_{t=w+1}^T
|\mathcal F_t(\outvar_t)-\mathcal F_{t-w}(\outvar_t)|
\leq\frac1w\sum_{t=w+1}^T\sum_{j=t-w}^{t-1}\delta_j
\leq V_{1,T},
\]
because each $\delta_j$ occurs at most $w$ times. Hence
$S\leq2Q+V_{1,T}\leq2TQ/w+V_{1,T}$, where the last inequality uses
$w\leq T$.
\end{proof}

Our next theorem provides an upper bound on the bilevel local regret of Algorithm~\ref{alg:online_functional_bo}.
\begin{theorem}\label{thrm:gd_bilevel_oracle_appdx}
Under Section~\ref{appx:FBO_assumptions_oracle}, Algorithm~\ref{alg:online_functional_bo} with $\alpha=1/L$ satisfies
\[
\mathbb E[\textBLR_w(T)]
\leq2L\left(\frac{2TQ}{w}+V_{1,T}\right)
+\frac{2T\sigma_f^2}{w}+2D_{T,w}.
\]
\end{theorem}

\begin{proof}
Set
\[
g_t:=\outgradtw(\outvar_t),\qquad
\widetilde g_t:=\outgradtildt(\outvar_t),\qquad
e_t:=\widetilde g_t-g_t.
\]
Since $\mathcal F_{t,w}$ is $L$-smooth and
$\outvar_{t+1}=\outvar_t-\alpha\widetilde g_t$,
\begin{align*}
\mathcal F_{t,w}(\outvar_{t+1})-\mathcal F_{t,w}(\outvar_t)
&\leq-\alpha\langle g_t,\widetilde g_t\rangle
+\frac{L\alpha^2}{2}\|\widetilde g_t\|^2.
\end{align*}
At $\alpha=1/L$, the identity
$2\langle g_t,\widetilde g_t\rangle
=\|g_t\|^2+\|\widetilde g_t\|^2-\|e_t\|^2$
yields
\[
\mathcal F_{t,w}(\outvar_{t+1})-\mathcal F_{t,w}(\outvar_t)
\leq-\frac1{2L}\|g_t\|^2+\frac1{2L}\|e_t\|^2.
\]
Summing, taking expectation over the full algorithmic history, and applying
Lemma~\ref{lem:stochastic_bounded_time_varying} gives
\[
\mathbb E[\textBLR_w(T)]
\leq2L\left(\frac{2TQ}{w}+V_{1,T}\right)
+\sum_{t=1}^T\mathbb E\|e_t\|^2.
\]
Finally, summing \eqref{eq:oracle_tracking_error} over $t$ gives
\[
\sum_{t=1}^T\mathbb E\|e_t\|^2
\leq\frac{2T\sigma_f^2}{w}+2D_{T,w},
\]
which proves the claim.
\end{proof}

\subsection{Functional Hypergradient}\label{appx:functional_hypergrad_algo}
In functional implicit differentiation, the function \textbf{\texttt{InnerOpt}} (defined in \cref{alg:inner_level}) optimizes inner model parameters for a given $\outvar$, initialization $\inNNparam_0$, and data $\inDataset$, using $\nbItersInnerSol$ gradient updates. It returns the inner model $\hat{\innerpred}_{\outvar}$, usually a neural network, parameterized by parameters $\inNNparam_\nbItersInnerSol$, approximating the inner-level solution. Similarly, \textbf{\texttt{AdjointOpt}} (defined in \cref{alg:adj_level}) optimizes adjoint model parameters with $\nbItersAdjSol$ gradient updates, producing the approximate adjoint function 
$\hat{\adjoint}_{\outvar}$. Other optimization procedures may also be used, especially when closed-form solutions are available, as exploited in the experiments in \cref{sec:experiments}. Operations requiring differentiation can be implemented using standard optimization procedures with automatic differentiation packages like PyTorch~\citep{PyTorch2019} or Jax~\citep{jax2018github}.

\vspace*{-0.3cm}
\begin{center}
\begin{minipage}{0.49\linewidth}
\begin{algorithm}[H]
  \caption{{ ~\bf\texttt{InnerOpt}} \phantom{$\hat{\innerpred}_{\outvar}$} }
  \label{alg:inner_level}
\begin{algorithmic}
    \Require outer variable $\outvar$, inner model $\innerpred$ parameterized by $\inNNparam_0$, dataset $\inDataset$
        \For{$m=0,\ldots,\nbItersInnerSol-1$}
            \State Sample batch ${\mathcal B}_{in}$ from $\inDataset$
            \State $\innerpred_{m} \leftarrow $ inner model parameterized by $\inNNparam_{m}$
            \State $g_{in} \!\leftarrow\! \nabla_{\inNNparam}[ \empinOBJ\parens{\outvar,\innerpred_{m},\inBatch} \!+\!  R_{in}(\inNNparam_{m})]$
            \State $\theta_{m+1}\leftarrow$ Update $\inNNparam_{m}$ using $g_{in}$
        \EndFor
        \State $\hat{\innerpred}_{\outvar} \leftarrow $ inner model parameterized by $\inNNparam_{m+1}$
        \State {\bf Return} $\hat{\innerpred}_{\outvar}$
\end{algorithmic}
\end{algorithm}
\end{minipage}
\hfill
\begin{minipage}{0.49\linewidth}
\begin{algorithm}[H]
  \caption{{~\bf \texttt{AdjointOpt}}}
  \label{alg:adj_level}
\begin{algorithmic}
    \Require outer variable $\outvar$, adjoint model $\adjoint$ parameterized by $\adjNNparam_0$, inner model $\hat{\innerpred}_{\outvar}$, dataset $\mathcal{D}$
        \For{$k=0,\ldots,\nbItersAdjSol-1$}
            \State \!\!\! \!\!\!\! Sample batch ${\mathcal B}$ from $\mathcal D$
            \State $\adjoint_{k} \leftarrow $ inner model parameterized by $\adjNNparam_{k}$
            \State \!\!\!$g_{adj} \!\leftarrow\! \nabla_{\adjNNparam} [ \empadjOBJ(\outvar, \adjoint_{k},\hat{\innerpred}_{\outvar},\mathcal{B})\!+\! R_{adj}(\adjNNparam_{k})]$
            \State \!\!\!$\adjNNparam_{k+1}\leftarrow$ Update $\adjNNparam_{k}$ using $g_{adj}$
        \EndFor
        \State $\hat{\adjoint}_{\outvar} \leftarrow $ adjoint model parameterized by $\adjNNparam_{m+1}$
        \State {\bf Return} $\hat{\adjoint}_{\outvar}$
\end{algorithmic}
\end{algorithm}
\end{minipage}
\end{center}

Since we cannot perform first-order optimization techniques directly in function spaces, we assume that, in practice, $\innerpred$ and $\adjoint$ are models parameterized by some finite dimensional parameter vectors $\inNNparam$ and $\adjNNparam$, rather than functions in $L_2$. As discussed in the theoretical analysis of the algorithm, we assume that these models map finite dimensional parameter vectors to a functions that are $\epsilon$-close to the true predictions functions. Together with empirical objectives, commonly used regularization techniques $R_{in}(\inNNparam)$ and $R_{adj}(\adjNNparam)$ may be introduced in inner and adjoint optimization subroutines, such as ridge penalty.

We approximate the hypergradient~$\outgradt$ after computing the approximate solutions $\hat{\innerpred}_{\outvar}$ and $\hat{\adjoint}_{\outvar}$. We decompose the gradient into two terms: $g_{Exp}$, an empirical approximation of $g_{\outvar}:=\partial_\outvar \outOBJ(\outvar, \innerpred^\star_{\outvar})$ representing the explicit dependence of $\outOBJ$ on the outer variable $\outvar$, and $g_{Imp}$, an approximation to the implicit gradient term $\crossDeriv_{\outvar}\adjoint_{\outvar}^{\star}$. Both terms are obtained by replacing the expectations by empirical averages over batches $\inBatch$ and $\outBatch$, and using the approximations $\hat{\innerpred}_{\outvar}$ and $\hat{\adjoint}_{\outvar}$ instead of the exact solutions. 

\subsection{Assumptions for Algorithm \ref{alg:smooth_fbo_stochastic} Convergence}\label{appx:SmoothFBO_assumptions}
In addition to the bounded-objective, variation, window, and domain conditions of Section~\ref{appx:FBO_assumptions_oracle}, assume:
\begin{enumerate}
    \item \textbf{FBO regularity}: At every $t$, all hypotheses of Lemmas E.4--E.5 in \citet{petrulionyte2024functional} hold, including the required strong-convexity, bounded-derivative, integrability, and smoothness conditions, with constants uniform in $t$. Consequently, each population outer objective $\mathcal F_t$ is differentiable and $\mathcal F_{t,w}$ has an $L$-Lipschitz gradient on $\Lambda$, with $L$ independent of the horizon.
    \item \textbf{Estimation-error decomposition}: Let $\mathcal A_{t-1}$ contain the information used to select $\outvar_t$ and the approximate inner and adjoint functions before the final hypergradient mini-batch, which is sampled conditionally independently given $\mathcal A_{t-1}$. Define
    \begin{align*}
    G_t&:=\mathbb E[\outgradhatt(\outvar_t)\mid\mathcal A_{t-1}],\\
    b_t&:=G_t-\outgradt(\outvar_t),\qquad
    \zeta_t:=\outgradhatt(\outvar_t)-G_t.
    \end{align*}
    Then $\zeta_t$ is $\mathcal A_t$-measurable,
    $\mathbb E[\zeta_t\mid\mathcal A_{t-1}]=0$, and
    $\mathbb E\|\zeta_t\|^2\leq\sigma_{\mathcal F_t}^2$.
    Set $\sigma_{\mathcal F}^2:=\sup_{t\geq1}
    \sigma_{\mathcal F_t}^2<\infty$.
    \item \textbf{Approximation bias}: With
    $B_T:=\sum_{t=1}^T\mathbb E\|b_t\|^2$, the convergence bound holds for any finite $B_T$. Lemma~\ref{lemma:bias} gives the specialization
    \[
    B_T\leq c_1\sum_{t=1}^T\epsilon_{\mathrm{in},t}
    +c_2\sum_{t=1}^T\epsilon_{\mathrm{adj},t}.
    \]
    Here $\epsilon_{\mathrm{in},t}$ and $\epsilon_{\mathrm{adj},t}$ are the round-$t$ expected inner- and adjoint-objective suboptimality tolerances in the assumptions of \citet{petrulionyte2024functional}.
    Sublinear regret uses $B_T=o(T)$ and $D_{T,w}=o(T)$.
\end{enumerate}
For $s<1$, all stored-gradient, bias, noise, and tracking summands are defined to be zero.
For Lemma~\ref{lem:linear_inner} only, additionally assume the
historical-point rate in \eqref{eq:historical_point_premise}.
The squared-error conditions stated in the main text are sufficient for
$B_T=o(T)$: by Cauchy--Schwarz,
\[
\sum_{t=1}^T\epsilon_{\mathrm{in},t}
\leq\sqrt{T\sum_{t=1}^T\epsilon_{\mathrm{in},t}^2}=o(T),
\]
and likewise for the adjoint tolerances.

\subsection{Convergence Analysis of Algorithm \ref{alg:smooth_fbo_stochastic}}

\subsubsection{Preliminary Lemmas}
We use the two following lemmas proven in \citet{petrulionyte2024functional} for bias-variance decomposition.

\begin{lemma}[Lemma E.4 in \citet{petrulionyte2024functional}]\label{lemma:bias}
Under Section~\ref{appx:SmoothFBO_assumptions}, the conditional bias $b_t=G_t-\outgradt(\outvar_t)$ satisfies
\[
\mathbb E\|b_t\|^2
\leq c_1\epsilon_{\mathrm{in},t}
+c_2\epsilon_{\mathrm{adj},t},
\]
where the horizon-independent constants $c_1,c_2\geq0$ are defined in Equation~50 of
\citet{petrulionyte2024functional}.
\end{lemma}

\begin{lemma}[Lemma E.5 in \citet{petrulionyte2024functional}]\label{lemma:variance}
Under the same assumptions, the centered error
$\zeta_t=\outgradhatt(\outvar_t)-G_t$ satisfies
\[
\mathbb E\|\zeta_t\|^2\leq\sigma_{\mathcal F_t}^2,
\]
where
\begin{align}\label{eq:variance_upper_bound}
\sigma^2_{\mathcal{F}_t} := \frac{2}{|\Bcal_{\text{out}}|} \left( 2 c_3^2 \mu^{-1} \epsilon_{\text{in},t} + \sigma_{\text{out}}^2 \right) + \frac{4 B_2^2}{|\Bcal_{\text{in}}|} \left( \mu^{-1} \epsilon_{\text{adj},t} + 2 \mu^{-3} c_4^2 M^2 \epsilon_{\text{in},t} + \mu^{-2} B_3^2 \right).
\end{align}
The remaining constants are defined in \citet{petrulionyte2024functional}.
\end{lemma}

\begin{lemma}[Expected Squared Error of Time-Smoothed Hypergradient Estimator]\label{lem:time_smoothed_error_apx}
Under Section~\ref{appx:SmoothFBO_assumptions},
\begin{align}
\mathbb E\!\left[\sum_{t=1}^T
\|\outgradtildt(\outvar_t)-\outgradtw(\outvar_t)\|^2\right]
\leq\frac{3T\sigma_{\mathcal F}^2}{w}+3B_T+3D_{T,w}.
\end{align}
In particular, Lemma~\ref{lemma:bias} gives
\[
B_T\leq c_1\sum_{t=1}^T\epsilon_{\mathrm{in},t}
+c_2\sum_{t=1}^T\epsilon_{\mathrm{adj},t}.
\]
Thus Lemma~\ref{lem:time_smoothed_error} holds with
$C_1=3$, $C_2=3c_1$, and $C_3=3c_2$, interpreting
$\sigma_{\mathcal F_t}^2$ there as the uniform bound
$\sigma_{\mathcal F}^2$.
\end{lemma}

\begin{proof}
By the definitions of $b_t$ and $\zeta_t$ and by adding and subtracting the historical exact hypergradients,
\begin{align*}
\outgradtildt(\outvar_t)-\outgradtw(\outvar_t)
&=\underbrace{\frac1w\sum_{i=0}^{w-1}\zeta_{t-i}}_{\bar\zeta_{t,w}}
+\underbrace{\frac1w\sum_{i=0}^{w-1}b_{t-i}}_{\bar b_{t,w}}
+r_{t,w}.
\end{align*}
This also proves the identity in Lemma~\ref{lem:time_smoothed}. The martingale-difference argument used in the proof of Lemma~\ref{lem:time_smoothed_oracle} gives
\[
\mathbb E\|\bar\zeta_{t,w}\|^2
=\frac1{w^2}\sum_{i=0}^{w-1}\mathbb E\|\zeta_{t-i}\|^2
\leq\frac{\sigma_{\mathcal F}^2}{w}.
\]
Jensen's inequality and zero padding give
\begin{align*}
\sum_{t=1}^T\mathbb E\|\bar b_{t,w}\|^2
&\leq\frac1w\sum_{t=1}^T\sum_{i=0}^{w-1}
\mathbb E\|b_{t-i}\|^2
\leq B_T,
\end{align*}
because each $b_s$, $1\leq s\leq T$, occurs in at most $w$ windows.
Finally, $\|a+b+c\|^2\leq3\|a\|^2+3\|b\|^2+3\|c\|^2$ yields
\begin{align*}
\mathbb E\!\left[\sum_{t=1}^T
\|\outgradtildt(\outvar_t)-\outgradtw(\outvar_t)\|^2\right]
&\leq3\sum_{t=1}^T\mathbb E\|\bar\zeta_{t,w}\|^2
+3\sum_{t=1}^T\mathbb E\|\bar b_{t,w}\|^2
+3D_{T,w}\\
&\leq\frac{3T\sigma_{\mathcal F}^2}{w}+3B_T+3D_{T,w}.
\end{align*}
Summing Lemma~\ref{lemma:bias} proves the specialization.
\end{proof}

\subsubsection{Convergence Theorem}
\begin{theorem}\label{thrm:alg1_rate_appdx}
Under Section~\ref{appx:SmoothFBO_assumptions}, Algorithm~\ref{alg:smooth_fbo_stochastic} with $\alpha=4/(5L)$ satisfies
\begin{align}
\mathbb E[\textBLR_w(T)]
&\leq\frac{5L}{2}\left(\frac{2TQ}{w}+V_{1,T}\right)
+\mathbb E[\Gamma_{T,w}]\notag\\
&\leq\frac{5L}{2}\left(\frac{2TQ}{w}+V_{1,T}\right)
+\frac{3T\sigma_{\mathcal F}^2}{w}+3B_T+3D_{T,w}.
\end{align}
Thus the constants in Theorem~\ref{thrm:alg1_rate} are
$C_4=5L/2$ and $C_5=1$.
\end{theorem}

\begin{proof}
Let
\[
g_t:=\outgradtw(\outvar_t),\qquad
\widetilde g_t:=\outgradtildt(\outvar_t),\qquad
e_t:=\widetilde g_t-g_t.
\]
By $L$-smoothness and the update
$\outvar_{t+1}=\outvar_t-\alpha\widetilde g_t$,
\[
\mathcal F_{t,w}(\outvar_{t+1})-\mathcal F_{t,w}(\outvar_t)
\leq-\alpha\langle g_t,\widetilde g_t\rangle
+\frac{L\alpha^2}{2}\|\widetilde g_t\|^2.
\]
Using
$2\langle g_t,\widetilde g_t\rangle
=\|g_t\|^2+\|\widetilde g_t\|^2-\|e_t\|^2$ gives
\begin{align*}
\mathcal F_{t,w}(\outvar_{t+1})-\mathcal F_{t,w}(\outvar_t)
&\leq-\frac\alpha2\|g_t\|^2
-\frac{\alpha(1-L\alpha)}2\|\widetilde g_t\|^2
+\frac\alpha2\|e_t\|^2.
\end{align*}
For every $0<\alpha\leq1/L$, the middle term is nonpositive, and hence
\[
\|g_t\|^2
\leq\frac2\alpha\left(
\mathcal F_{t,w}(\outvar_t)-\mathcal F_{t,w}(\outvar_{t+1})
\right)+\|e_t\|^2.
\]
Summing, taking expectation over the full algorithmic history, applying
Lemma~\ref{lem:stochastic_bounded_time_varying}, and setting
$\alpha=4/(5L)$ yields
\[
\mathbb E[\textBLR_w(T)]
\leq\frac{5L}{2}\left(\frac{2TQ}{w}+V_{1,T}\right)
+\sum_{t=1}^T\mathbb E\|e_t\|^2.
\]
Lemma~\ref{lem:time_smoothed_error_apx} completes the proof.
\end{proof}

\subsection{Reduction of Rates in Linear Predictor Setting}

\begin{lemma}[Reduction of Rates with Linear Inner Predictor]
Suppose
$\innerpred^\star_{t,\outvar}(x)=\Phi(x)\theta^\star_{t,\outvar}$,
and assume the corresponding parametric historical-point local regret obeys the rate in \citet{lin2023non,bohne2024online}. Then
\[
\mathbb E[\textBLR_w(T)]
\leq\bigO\left(
\frac{TQ}{w}+V_{1,T}
+\frac{T\sigma_{\mathcal F}^2}{w}
+H_{2,T}+D_{T,w}\right),
\]
where
\[
H_{2,T}:=\sum_{t=2}^T\sup_{\boldsymbol\lambda\in\Lambda}
\|\theta^\star_{t-1,\boldsymbol\lambda}
-\theta^\star_{t,\boldsymbol\lambda}\|^2.
\]
\end{lemma}

\begin{proof}
The linear representation reduces \eqref{def:FBO} to a parametric bilevel problem over $\theta$. Define the historical-point window gradient
\[
h_{t,w}:=\frac1w\sum_{i=0}^{w-1}
\outgradti(\outvar_{t-i}).
\]
The required historical-point premise is
\begin{equation}\label{eq:historical_point_premise}
\sum_{t=1}^T\mathbb E\|h_{t,w}\|^2
\leq\bigO\left(
\frac{TQ}{w}+V_{1,T}
+\frac{T\sigma_{\mathcal F}^2}{w}+H_{2,T}\right).
\end{equation}
It does not identify $h_{t,w}$ with the common-current-point target in
Definition~\ref{def:bilevel_local_regret}. Instead, by the definition of
$r_{t,w}$,
\[
h_{t,w}=\outgradtw(\outvar_t)+r_{t,w}.
\]
Consequently,
\[
\|\outgradtw(\outvar_t)\|^2
=\|h_{t,w}-r_{t,w}\|^2
\leq2\|h_{t,w}\|^2+2\|r_{t,w}\|^2.
\]
Summing and using the definition of $D_{T,w}$ proves the claim. The stated
sublinear rate follows when $w=w_T\to\infty$, $w_T=o(T)$, and
$V_{1,T}$, $H_{2,T}$, and $D_{T,w}$ are $o(T)$.
\end{proof}

\subsection{Additional Lemmas}

\begin{lemma}[Approximate Gradient Norm Bound]\label{lem:approx_gradient_norm}
Let \(\outgradtildt(\outvar_t)\) be an approximate gradient and \(\outgradtw(\outvar_t)\) the true gradient of the loss \(\Ftw{\outvar_t}\) at iterate \(\outvar_t\). We have
\begin{align*}
-\left\| \outgradtildt(\outvar_t) \right\|^2 \leq -\frac{1}{2} \left\| \outgradtw(\outvar_t) \right\|^2 + \left\| \outgradtildt(\outvar_t) - \outgradtw(\outvar_t) \right\|^2.
\end{align*}
\end{lemma}
\begin{proof}
Consider the norm of the true gradient:
\begin{align*}
\left\| \outgradtw(\outvar_t) \right\|^2 \leq 2 \left\| \outgradtildt(\outvar_t) \right\|^2 + 2 \left\| \outgradtildt(\outvar_t) - \outgradtw(\outvar_t) \right\|^2.
\end{align*}
Rearrange:
\begin{align*}
-\frac{1}{2} \left\| \outgradtw(\outvar_t) \right\|^2 \geq -\left\| \outgradtildt(\outvar_t) \right\|^2 - \left\| \outgradtildt(\outvar_t) - \outgradtw(\outvar_t) \right\|^2.
\end{align*}
Add \(\left\| \outgradtildt(\outvar_t) - \outgradtw(\outvar_t) \right\|^2\) to both sides:
\begin{align*}
-\left\| \outgradtildt(\outvar_t) \right\|^2 \leq -\frac{1}{2} \left\| \outgradtw(\outvar_t) \right\|^2 + \left\| \outgradtildt(\outvar_t) - \outgradtw(\outvar_t) \right\|^2.
\end{align*}
\end{proof}

\begin{lemma}[Generalized Projection Inequality]\label{lem:stochastic_bregman_inequality}
Let \(\outgradtildt(\outvar_t)\) be an approximate gradient and \(\outgradtw(\outvar_t)\) the true gradient of the loss \(\Ftw{\outvar_t}\) at iterate \(\outvar_t\). We have
\begin{align*}
\left\langle \outgradtildt(\outvar_t) - \outgradtw(\outvar_t), \outgradtildt(\outvar_t) \right\rangle \leq \left\| \outgradtildt(\outvar_t) - \outgradtw(\outvar_t) \right\|^2 + \frac{1}{4} \left\| \outgradtildt(\outvar_t) \right\|^2.
\end{align*}
\end{lemma}
\begin{proof}
Consider the inner product:
\begin{align*}
\left\langle \outgradtildt(\outvar_t) - \outgradtw(\outvar_t), \outgradtildt(\outvar_t) \right\rangle.
\end{align*}
By Young's inequality, for any \(\eta > 0\):
\begin{align*}
\left\langle \mathbf{a}, \mathbf{b} \right\rangle \leq \frac{\eta}{2} \left\| \mathbf{a} \right\|^2 + \frac{1}{2\eta} \left\| \mathbf{b} \right\|^2.
\end{align*}
Set \(\mathbf{a} = \outgradtildt(\outvar_t) - \outgradtw(\outvar_t)\), \(\mathbf{b} = \outgradtildt(\outvar_t)\), and \(\eta = 2\):
\begin{align*}
\left\langle \outgradtildt(\outvar_t) - \outgradtw(\outvar_t), \outgradtildt(\outvar_t) \right\rangle &\leq \left\| \outgradtildt(\outvar_t) - \outgradtw(\outvar_t) \right\|^2 \\
&\quad + \frac{1}{4} \left\| \outgradtildt(\outvar_t) \right\|^2.
\end{align*}
\end{proof}

\section{Additional Experimental Details}\label{appx:experiment}

The empirical curves labeled as BLR in the implementation use the stored-gradient proxy
\[
\widehat{\mathrm{BLR}}_w(T):=\sum_{t=1}^T
\|\outgradtildt(\outvar_t)\|^2,
\]
rather than the generally unobserved population quantity in Definition~\ref{def:bilevel_local_regret}. Since
$\outgradtildt(\outvar_t)=\outgradtw(\outvar_t)+e_t$,
\begin{equation}\label{eq:empirical_blr_proxy}
\mathbb E[\widehat{\mathrm{BLR}}_w(T)]
\leq2\mathbb E[\mathrm{BLR}_w(T)]+2\mathbb E[\Gamma_{T,w}].
\end{equation}
Thus the plots diagnose the stored gradients used for the updates, but do not by themselves verify the theorem's asymptotic conditions.

\subsection{Non\mbox{-}stationary Regression}

\begin{figure}[H]
    \centering
    \includegraphics[width=0.45\linewidth]{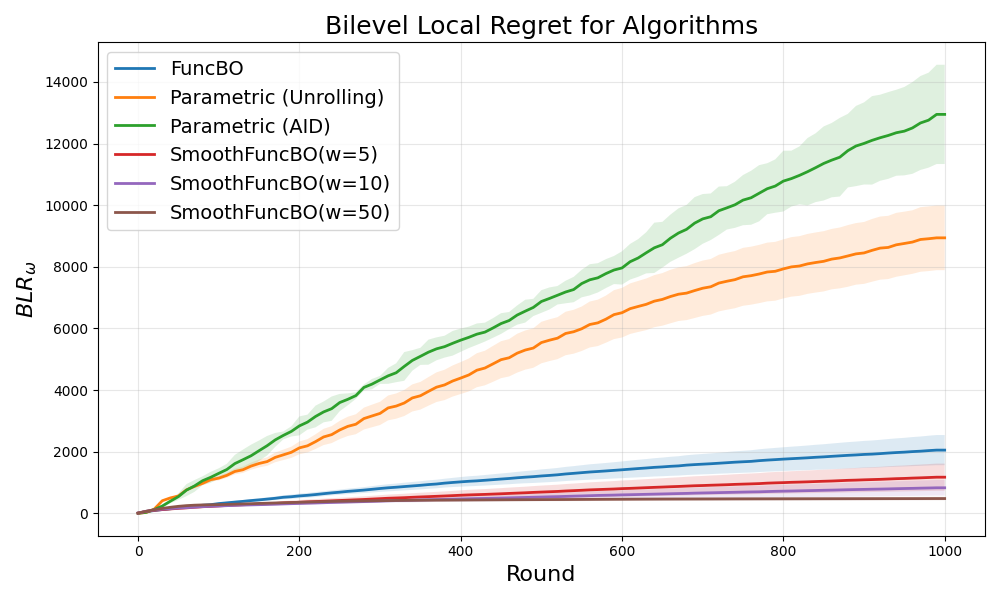}
    \includegraphics[width=0.45\linewidth]{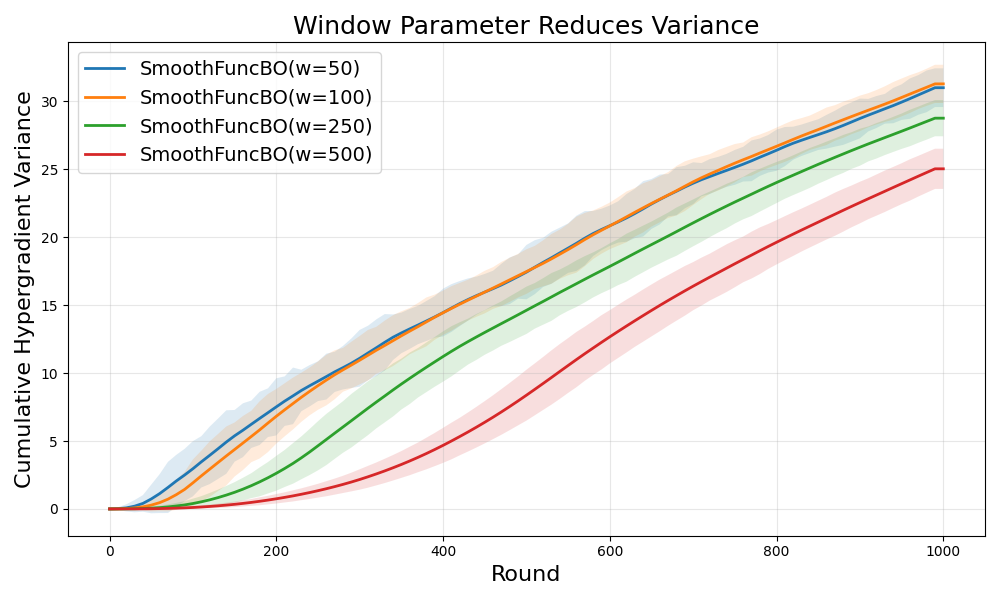}
    \vspace{-6pt}
    \caption{\textbf{Effect of smoothing on the stored-gradient proxy and hypergradient variance.}
    (Left) Cumulative stored-gradient BLR proxy on the sinusoidal-drift task; curves are averaged over seeds.
    (Right) Variance of the hypergradient as a function of the smoothing window \(w\); increasing \(w\) reduces variance. Shaded regions indicate \(95\%\) confidence intervals across seeds.}
    \label{fig:blr_comparison}
\end{figure}

We analyze how the smoothing window \(w\) affects hypergradient variance and the stored-gradient BLR proxy. Figure~\ref{fig:blr_comparison} summarizes these effects: the left panel reports the cumulative proxy and the right panel shows that larger \(w\) reduces the variance of the outer hypergradient. Over the plotted horizon, temporal smoothing yields smoother updates and lower proxy values.

\begin{figure}[H]
    \centering
    \includegraphics[width=0.45\linewidth]{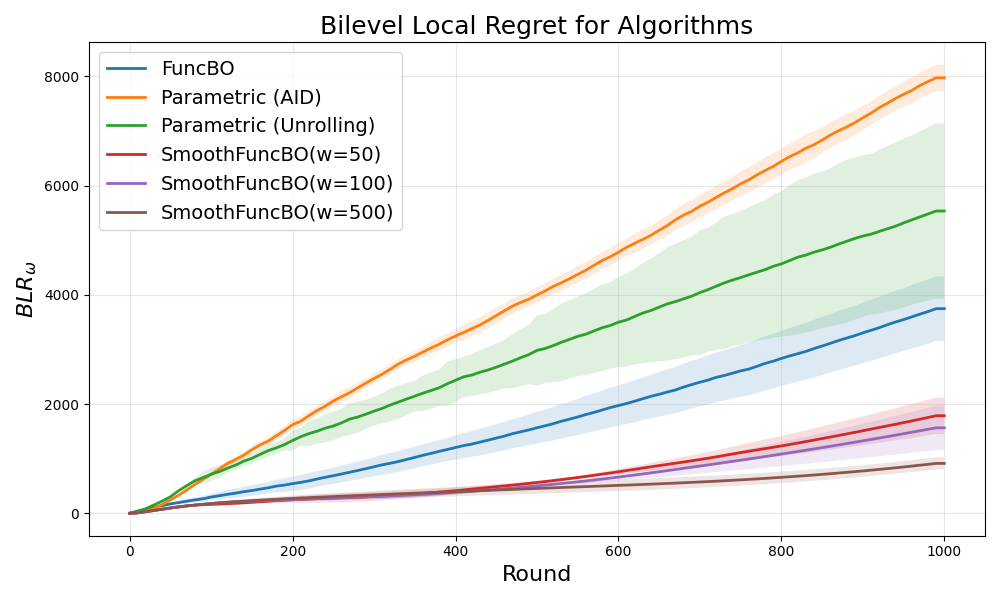}\includegraphics[width=0.45\linewidth]{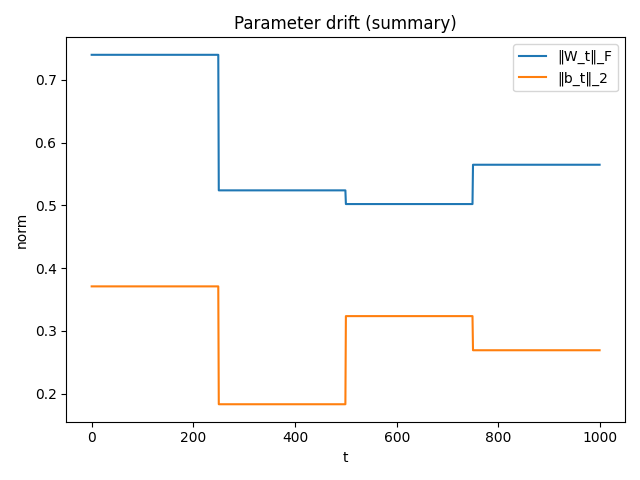}
    \caption{\textbf{Behavior under discrete jump–based non-stationarity.}
    (Left) Cumulative stored-gradient BLR proxy under abrupt parameter jumps.
    (Right) Evolution of \((W_t,b_t)\) with discrete jumps that induce the drift. \textbf{SmoothFBO} recovers quickly after each jump and maintains lower cumulative regret.}
    \label{fig:variance_regret}
\end{figure}

To complement the sinusoidal drift analyzed in the main text (Fig.~\ref{fig:drift}), we also examine a \emph{discrete jump–based} non-stationarity where \((W_t,b_t)\) undergo abrupt changes at fixed intervals. Figure~\ref{fig:variance_regret} reports the stored-gradient proxy under jump-based drift and visualizes the parameter trajectory. \textbf{SmoothFBO} attains substantially lower proxy values than \textbf{FBO} and parametric baselines (AID, Unrolling) over the plotted horizon.

\paragraph{Hyperparameter selection.}
All results are averaged over three random seeds, with shaded regions indicating standard error. Our method demonstrates robust performance across a broad range of hyperparameter configurations. Unless otherwise stated, reported results use an inner learning rate of $10^{-4}$, an outer learning rate of $10^{-3}$, a batch size of $32$, and $5$ inner steps. We observe consistent \textbf{SmoothFBO} performance across the following ranges:

\begin{itemize}
\item Inner learning rate: $\{10^{-2}, 10^{-3}, 10^{-4}\}$.
\item Outer learning rate: $\{10^{-3}, 10^{-2}\}$.
\item Batch size: $\{16, 32, 64, 128\}$.
\item Number of inner steps: $\{5, 10\}$.
\end{itemize}

Training is performed for $1000$ outer steps while varying the window parameter, which controls the time smoothing of hypergradients, over $\{1, 5, 10, 50, 100, 250, 500\}$. In the \textbf{SmoothFBO} implementation, a gradient buffer maintains recent hypergradients to enable this smoothing mechanism. Stable performance across these settings highlights the method’s robustness to hyperparameter choice.

\subsection{Reinforcement Learning}

\paragraph{Non-stationary CartPole environment.}  
As described in the main text, our non-stationary variant of CartPole modifies the reward interval associated with the pole angle. In the stationary environment, the reward is $1$ when the pole angle lies in a fixed optimal interval and $0$ otherwise. In the non-stationary setting, this interval shifts gradually throughout training. Formally, we interpolate linearly between two distinct pole-angle intervals: 
\[
(-0.2095,\, 0.06) \quad \longrightarrow \quad (-0.06,\, 0.2095).
\] 
This interpolation induces a continuous drift in the reward boundaries, requiring the agent to track the changing environment. The main text (Fig.~\ref{fig:non_s_reward}) visualizes the pole angle drift for different agents.

\begin{figure}[H]
    \centering
    \includegraphics[width=0.45\linewidth]{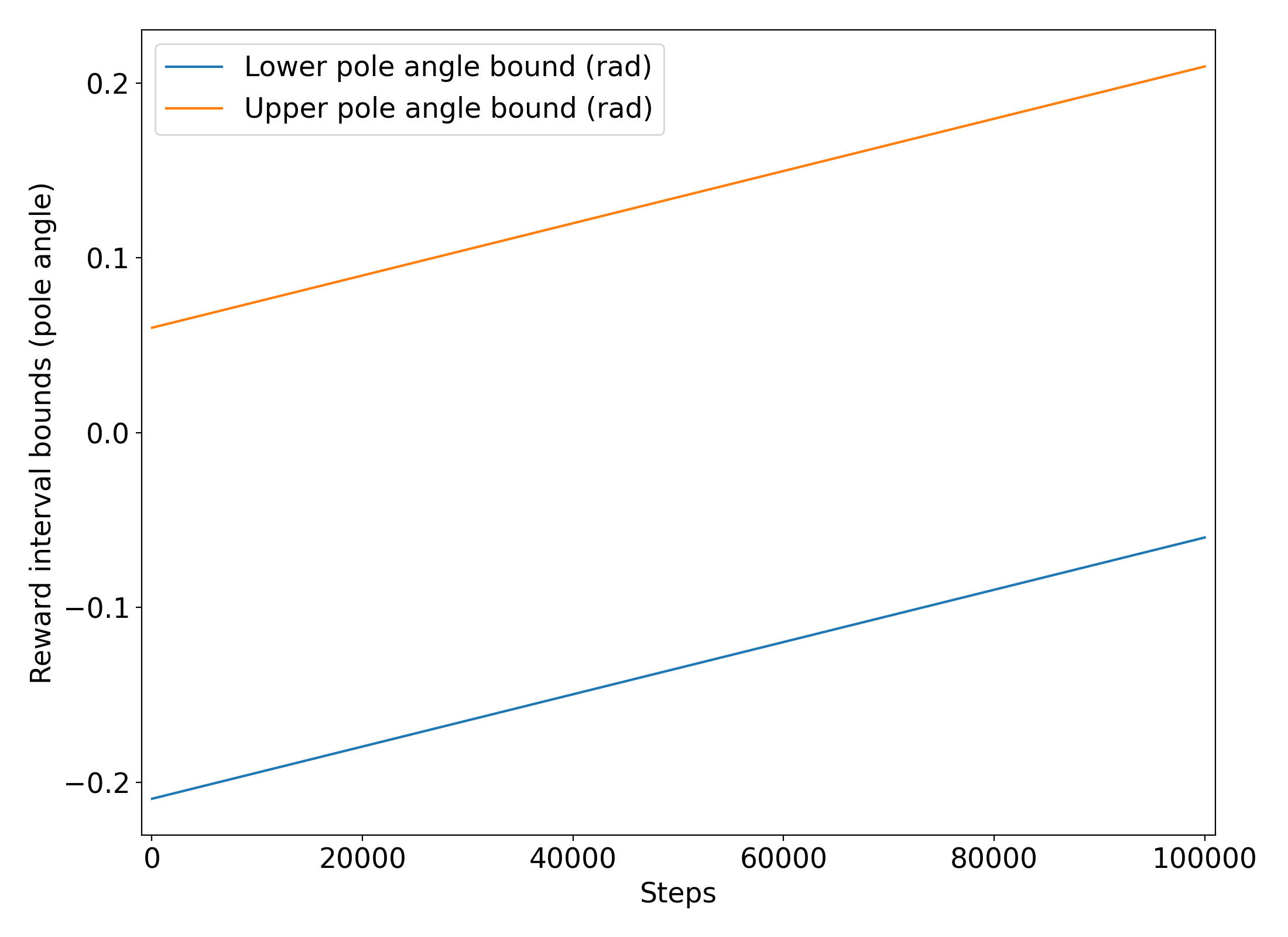}
    \caption{Non-stationary CartPole reward interval. This illustrates how the reward dynamics shift over training steps. The agent is then forced to update its policy accordingly as illustrated in Fig.~\ref{fig:non_s_reward}.}
    \label{fig:non_s_reward_bounds}
\end{figure}

\paragraph{Implementation details.}  
Experiments were implemented in PyTorch using the OpenAI Gym CartPole environment \citep{brockman2016openai}.  
Runs were executed on \textit{24GB NVIDIA RTX A5000} GPUs.  
A single configuration requires approximately $8$ hours to complete one million environment steps. We used Adam optimizers throughout.

\subsubsection{Additional Results}

\begin{figure}[H]
    \centering
    \includegraphics[width=0.45\linewidth]{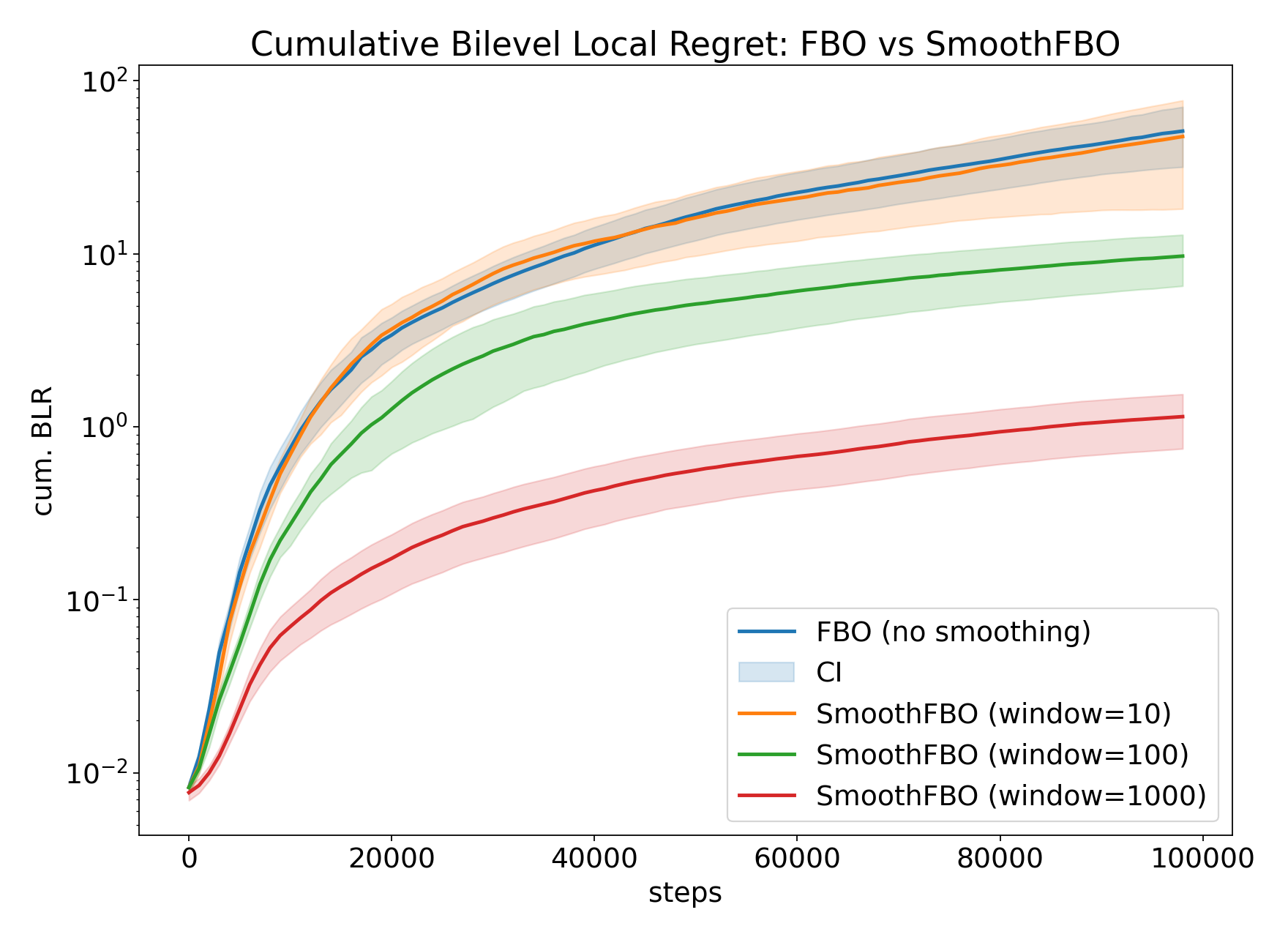}\includegraphics[width=0.45\linewidth]{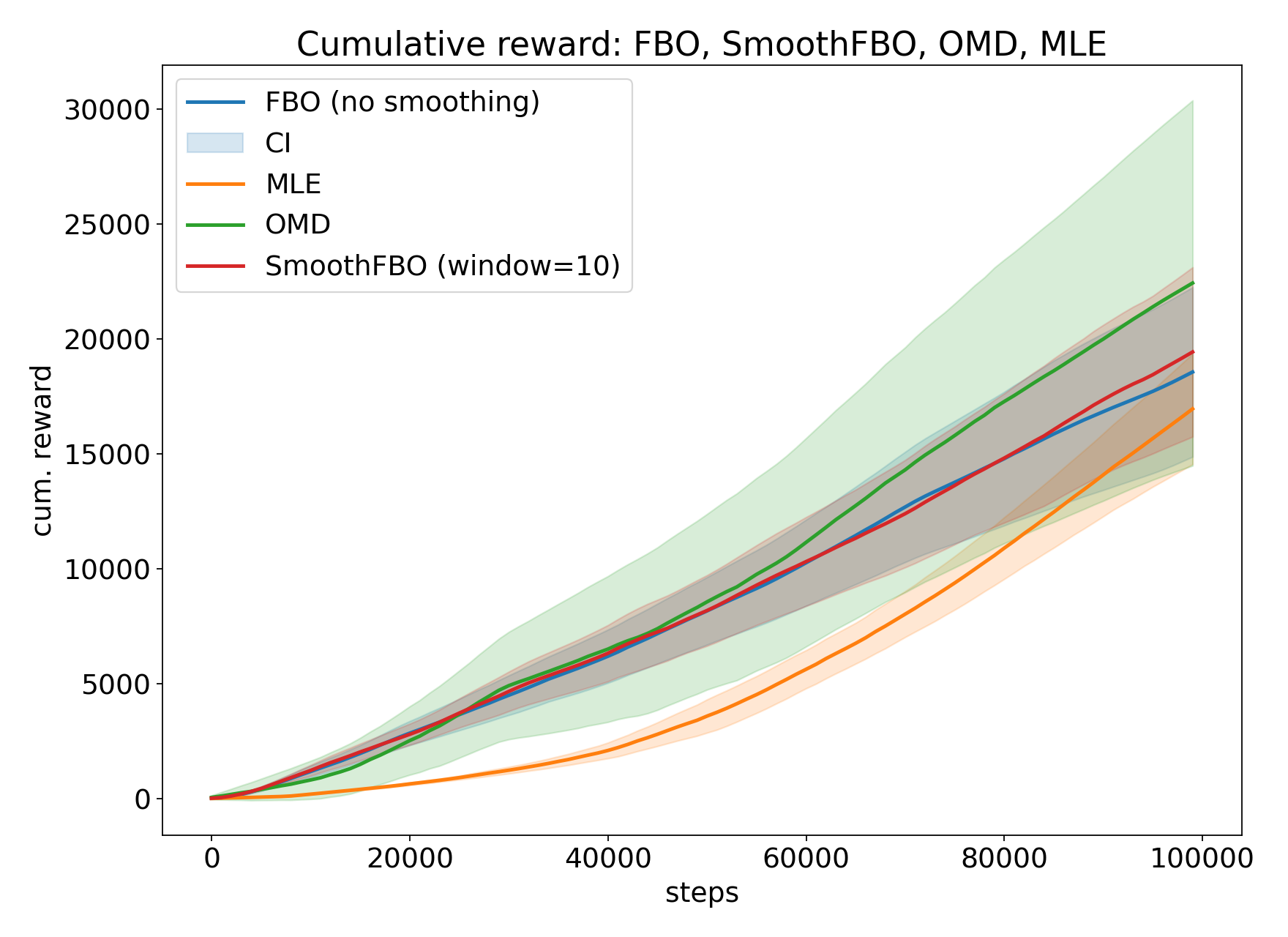}
    \caption{\textbf{Effect of hypergradient smoothing.} 
    (Left) Stored-gradient BLR proxy in the training environment for \textbf{SmoothFBO} with varying window $w$ versus \textbf{FBO} (no smoothing); larger $w$ lowers the proxy and reduces outer-update variability. 
    (Right) Cumulative reward in the evaluation environment for \textbf{SmoothFBO} ($w{=}10$) compared to \textbf{FBO}, \textbf{OMD}, and \textbf{MLE}. The results are averaged over 20 seeds. Shaded regions show $95\%$ confidence intervals.}
    \label{fig:rl_appendix_smoothing}
\end{figure}

Figure~\ref{fig:rl_appendix_smoothing} complements the main RL results by isolating the effect of hypergradient time-smoothing on (left) the stored-gradient BLR proxy and (right) cumulative reward over $100$k environment steps. Increasing the smoothing window $w$ lowers the proxy and reduces variability in the outer updates over the plotted horizon.

\subsubsection{Algorithm Configuration}

For all three methods described in \ref{sec:experiments}, we used neural networks with hidden dimensions of 3 for the world model, resulting in the under-specified setting as in \citet{Nikishin:2022}. We use soft Q-learning with temperature parameter $\alpha$ to encourage exploration. Each method was tuned in the stationary environment to ensure fair comparison, with the same hyperparameters used for the non-stationary evaluation.

\paragraph{Hyperparameter selection.} We conduct a grid search in the stationary scenario using seed $1$ for all three methods with the following values:
\begin{itemize}
    \item Inner learning rate: $\{3 \cdot 10^{-3}, 3 \cdot 10^{-4}\}$
    \item Parameter $\tau$: $\{10^{-1}, 10^{-2}, 10^{-3}\}$
    \item Temperature parameter $\alpha$: $\{10^{-1}, 10^{-2}, 10^{-3}\}$
\end{itemize}
We use a replay buffer with capacity $100000$ and perform $1000000$ training steps during tuning. In the \textit{SmoothFBO} implementation, we introduce a gradient buffer that maintains a history of recent hypergradients. The gradient buffer size and the smoothing parameter $\theta$ control the level of temporal averaging. These parameters were tuned to balance adaptation speed and stability over $\theta=0.4,0.6,0.8$ and hypergradient buffer size of $1000, 10000$.

\end{document}